\documentclass[a4paper,fleqn]{cas-dc}

\usepackage[numbers]{natbib}
\usepackage{color}
\usepackage{hyperref}
\newcommand{\red}[1]{\textcolor{red}{#1}}

\def\tsc#1{\csdef{#1}{\textsc{\lowercase{#1}}\xspace}}
\tsc{WGM}
\tsc{QE}

\begin{document}
\let\WriteBookmarks\relax
\def\floatpagepagefraction{1}
\def\textpagefraction{.001}

\shorttitle{Decoupled Cross-Modal Alignment Network for Text-RGBT Person Retrieval and A High-Quality Benchmark}    

\shortauthors{Yifei Deng, Chenglong Li, Zhenyu Chen, Zihen Xu and Jin Tang}  

\title[mode = title]{Decoupled Cross-Modal Alignment Network for Text-RGBT Person Retrieval and A High-Quality Benchmark}  



%

\author[1,4]{Yifei Deng}[
      ]
\ead{yf-ah@foxmail.com}

\author[2,3]{Chenglong Li}[
      orcid=0000-0002-7233-2739,
      ]
\cormark[1]
\cortext[1]{Corresponding author.}
\ead{lcl1314@foxmail.com}
\author[2,4]{Zhengyu Chen}[
      ]
\ead{wa24301145@stu.ahu.edu.cn}
\author[2,4]{Zihen Xu}[
      ]
\ead{wa24201021@stu.ahu.edu.cn}
\author[1,4]{Jin Tang}[
      orcid=0000-0002-4123-268X,
      ]
\ead{tangjin@ahu.edu.cn}

\affiliation[1]{organization={
School of Computer Science and Technology, Anhui University},
city={Hefei},
postcode={230601}, 
state={Anhui},
country={China}
}
\affiliation[2]{organization={
School of Artificial Intelligence, Anhui University},
city={Hefei},
postcode={230601}, 
state={Anhui},
country={China}
}
\affiliation[3]{organization={
National Key Laboratory of Opto-Electronic Information Acquisition and Protection Technology},
city={Hefei},
postcode={230601}, 
state={Anhui},
country={China}
}
\affiliation[4]{organization={
Anhui Provincial Key Laboratory of Multimodal Cognitive Computation},
city={Hefei},
postcode={230601}, 
state={Anhui},
country={China}
}

\begin{abstract}
The performance of traditional text-image person retrieval task is easily affected by lighting variations due to imaging limitations of visible spectrum sensors. 
In recent years, cross-modal information fusion has emerged as an effective strategy to enhance retrieval robustness. By integrating complementary information from different spectral modalities, it becomes possible to achieve more stable person recognition and matching under complex real-world conditions.
Motivated by this, we introduce a novel task: Text-RGBT Person Retrieval, which incorporates cross-spectrum information fusion by combining the complementary cues from visible and thermal modalities for robust person retrieval in challenging environments.
The key challenge of Text-RGBT person retrieval lies in aligning text with multi-modal visual features.
However, the inherent heterogeneity between visible and thermal modalities may interfere with the alignment between vision and language.
To handle this problem, we propose a Decoupled Cross-modal Alignment network (DCAlign), which sufficiently mines the  relationships between modality-specific and modality-collaborative visual with the text, for Text-RGBT person retrieval. 
To promote the research and development of this field, we create a high-quality Text-RGBT person retrieval dataset, RGBT-PEDES.
RGBT-PEDES contains 1,822 identities from different age groups and genders with 4,723 pairs of calibrated RGB and T images, and covers high-diverse scenes from both daytime and nighttime with a various of challenges such as occlusion, weak alignment and adverse lighting conditions.
Additionally, we carefully annotate 7,987 fine-grained textual descriptions for all RGBT person image pairs.   
Extensive experiments on RGBT-PEDES demonstrate that our method outperforms existing text-image person retrieval methods.
The code and dataset will be released on \href{https://github.com/Yifei-AHU/RGBT-PEDE}{https://github.com/Yifei-AHU/RGBT-PEDE}.
\end{abstract}


\begin{highlights}
\item We are the first to define the Text-RGBT person retrieval task, extending text-based cross-modal retrieval to real-world scenarios with multi-modal visual inputs.
\item We propose a decoupled cross-modal alignment framework that structurally aligns text with RGBT visual representations via modality-collaborative and modality-specific visual-text alignment, enabling robust and fine-grained Text-RGBT retrieval.
\item We construct a high-quality Text-RGBT dataset (RGBT-PEDES) featuring diverse and challenging conditions, setting a new benchmark for future research in this field.
\end{highlights}


\begin{keywords}
Text-RGBT \sep Person Retrieval \sep Decoupled Cross-modal Alignment \sep Benchmark \sep RGBT-PEDES \sep
\end{keywords}

\maketitle

\section{Introduction}
\label{sec:intro}
Traditional text-image person retrieval is a task that matches person images in an image database based on captions~\cite{li2017person}~\cite{wang2019camp}~\cite{jiang2025attributes}~\cite{gong2024cross}~\cite{deng2025uncertainty}. 
This technology has broad application potential in areas such as social media analysis and video surveillance, providing significant value for smart city development and public safety management. 
With the advancement of deep learning~\cite{deng2024collaborative}~\cite{mohammed2023comprehensive}~\cite{deng2022factory}~\cite{wang2025action}~\cite{li2025adapting} and multimodal visual-language models~\cite{wang2024structural}~\cite{lin2024vila}~\cite{dai2025humanvlm}~\cite{quan2025multi}, text-image person retrieval has gradually garnered attention from researchers, resulting in a series of notable breakthroughs. 
However, current text-image person retrieval is highly sensitive to lighting variations, leading to unreliable results due to the imaging limitations of visible spectrum sensors, which can only capture visible light (0.38–0.78$\mu m$) reflected from objects.
Different from visible cameras, thermal imaging cameras can operate under various lighting conditions because they capture infrared radiation emitted by the surfaces of all objects with temperatures above absolute zero~\cite{ha2017mfnet}~\cite{zhou2022edge}~\cite{ring2012infrared}.
This motivates the use of cross-modal information fusion as a promising solution to overcome the limitations of single-modality 
input.
By integrating complementary cues from visible and thermal modalities, it becomes possible to enhance retrieval robustness and achieve more reliable person understanding in complex environments.

\begin{figure}[!t]
\centering
\includegraphics[width=3.2in,height=2.2in]{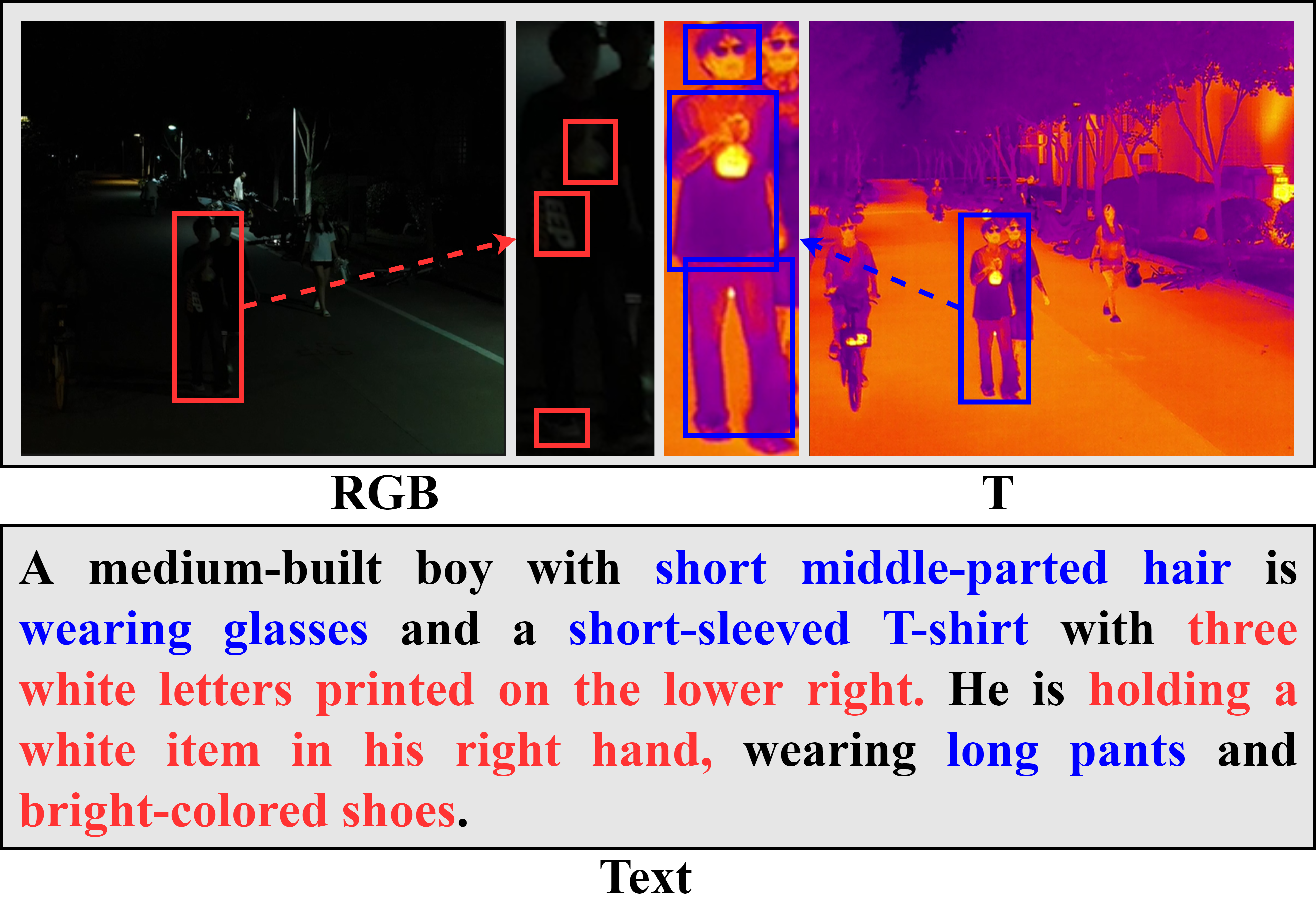}
\caption{Visualization of the RGB image, T image, and text: The red text corresponds to the region highlighted by the red bounding box in the RGB image, while the blue text refers to the region marked by the blue bounding box in the T image.}
\label{fig_1}
\end{figure}

To achieve robust and precise cross-modal person retrieval in complex scenarios, we propose a new task, called Text-RGBT person retrieval, which aims to retrieve paired visible and thermal person images using textual captions.
The core of this task lies in effective information fusion across modalities, fully leveraging the complementary characteristics of visible and thermal images to enhance semantic understanding and retrieval robustness under diverse conditions.
Figure~\ref{fig_1} illustrates several RGBT person image pairs along with their corresponding textual descriptions of pedestrian appearance. 
As shown, visible and thermal images provide complementary information, facilitating accurate person identification under challenging lighting conditions, such as low-light or strong-light environments.  
This new task opens up new possibilities for applying cross-modal text-image person retrieval in complex scenarios. 
To the best of our knowledge, this is the first study addressing the Text-RGBT person retrieval problem.

To effectively align text with multi-modal visual representations and fully exploit the complementary semantic perception capabilities of RGB and T modalities, we propose a Decoupled Cross-modal Alignment Network (DCAlign), which systematically integrates both modality-collaborative and modality-specific visual-text alignment strategies, aiming to enhance semantic matching between RGBT multi-modal representations and textual embeddings.
Specifically, we begin by analyzing the color dependency of semantic attributes, dividing textual attributes into three hierarchical levels: color-related, color-unrelated, and arbitrary attributes. 
Based on this categorization, we design a modality-specific text masking mechanism, including color-related attribute masks, color-unrelated attribute masks, and random masks. 
By selectively masking different semantic components in the text, this mechanism guides the model to explicitly learn the correspondence between semantic attributes and specific visual modalities during training, thereby enabling structured semantic modeling and attribute-aware alignment.

Building upon the semantic-level decomposition and masking strategy, DCAlign further constructs two complementary alignment strategies to align text with visual modalities more effectively:
First, the modality-collaborative visual-text alignment focuses on aligning the fused representations of RGB and T modalities with masked text containing arbitaray semantic attributes.
We introduce both identity-level semantic distribution matching and explicit token-level reconstruction mechanisms to construct fine-grained cross-modal semantic associations, thus improving the robustness and expressiveness of cross-modal matching.
Second, the modality-specific visual-text alignment targets the perceptual preferences of individual modalities.
On one hand, RGB images are aligned with color-related text via instance-level global alignment. On the other hand, T images are aligned with color-unrelated text through a unidirectional semantic binding mechanism, which helps preserve the modality’s intrinsic perception capability. 
Additionally, we incorporate a semantic consistency based soft local reconstruction loss to establish flexible and semantically adaptive fine-grained associations between vision and language, thereby fully leveraging the representational strengths of each modality across different semantic levels.

Moreover, we construct a high-quality Text-RGBT person retrieval dataset named RGBT-PEDES.  
This dataset includes 1,822 person identities, comprising a total of 4,723 pairs of person images that span both RGB and T, accompanied by 7,987 textual descriptions.
These descriptions were created by 23 annotators, each providing detailed accounts of the appearance of person based on both RGB and thermal modality images.  
To the best of our knowledge, this is the first dataset specifically designed for the task of Text-RGBT person retrieval.  
Compared to existing text-based person retrieval datasets, RGBT-PEDES introduces the thermal modality, expanding the retrieval applicability to more complex scenarios, making it reliable even under challenging lighting conditions.  
Additionally, the dataset presents a higher level of difficulty, covering diverse conditions such as low light, occlusion, abnormal lighting, darkness, glare, and weak alignment.

Extensive experiments on RGBT-PEDES demonstrate that DCAlign achieves superior performance compared to the latest text-image person retrieval method. 
The contributions of this paper can be summarized as follows:
\begin{itemize}
\item{We are the first to propose the task of Text-RGBT person retrieval, enabling text-based cross-modal person retrieval to be applicable in various complex scenarios.}
\item{We propose a decoupled cross-modal alignment framework that structurally aligns text with RGBT visual representations via modality-collaborative and modality-specific visual-text alignment, enabling robust and fine-grained Text-RGBT retrieval.}
\item{We contribute a high-quality Text-RGBT person retrieval dataset that captures a wide range of real-world challenges, offering a comprehensive and diverse benchmark to facilitate progress in cross-modal person retrieval research.}
\item{Extensive experiments conducted on the proposed RGBT-PEDES dataset demonstrate that our method outperforms the existing state-of-the-art text-image person retrieval methods}
\end{itemize}

\section{Related Work}
\subsection{Text-Image Person Retrieval}
The text-image person retrieval task is first introduced by \cite{li2017person} to retrieve target person images from a gallery based on natural language descriptions.
Early methods mainly focus on global feature alignment between images and texts, typically employing specific loss functions to achieve cross-modal matching, such as \cite{chen2021cross} and \cite{sarafianos2019adversarial}. However, these approaches fail to fully capture fine-grained semantic correspondences, limiting the potential for further improvement in retrieval accuracy.
To address this limitation, Chen et al.\cite{chen2018improving} propose a patch-word level matching framework; Ya et al.\cite{jing2020pose} incorporate pose priors to design a multi-granularity attention mechanism for enhanced local alignment; and Niu et al.\cite{niu2020improving} further refine the multi-granularity alignment strategy.
Although these approaches achieve significant performance gains, they often come at the cost of increased computational complexity and do not fully exploit the representational power of modern pre-trained vision-language models\cite{xu2022simple, zeng2023x}.
To this end, Han et al.\cite{han2021text} are the first to incorporate CLIP into the text-image person retrieval task. By leveraging momentum contrastive learning, they effectively transfer knowledge from large-scale image-text pairs to person-text matching, leading to a breakthrough in this field.
Building upon this, Jiang et al.\cite{jiang2023cross} further enhance fine-grained cross-modal matching by exploiting the visual and textual encoders of CLIP. Zuo et al. extend this direction by constructing a new benchmark for ultra-fine-grained text-image person retrieval and proposing an efficient fine-grained alignment approach based on a shared cross-modal granularity decoder. Meanwhile, Cao et al. systematically investigate the potential of vision-language models in downstream text-image person retrieval and propose TBPS-CLIP, a strong and representative baseline for this task.
However, most existing studies focus on well-lit and ideal environments, severely limiting their practical applicability in complex real-world scenarios.
To overcome this limitation, we propose the Text-RGBT person retrieval task, which leverages the advantages of thermal modalities under low-light conditions to enable robust person retrieval under arbitrary illumination settings.

\subsection{RGBT Multi-modal Learning}
The thermal modality gains increasing attention in the computer vision community due to its inherent insensitivity to lighting variations, which enables consistent performance in low-light or nighttime conditions. This unique property makes the fusion of RGB and thermal modalities particularly appealing for a wide range of visual tasks, including detection, tracking, segmentation, and recognition~\cite{feng2024rgbt, tu2022rgbt, liu2021cross, wang2024alignment}.
Motivated by these advantages, a growing number of studies integrate the thermal modality into multi-modal frameworks to enhance the robustness and adaptability of vision systems in complex or adverse environments.
As an early pioneer in this direction, CO Conaire et al.\cite{conaire2006comparison} propose a deep learning-based fusion approach that accurately tracks objects in cluttered scenes by leveraging complementary cues from RGB and thermal infrared streams.
Building on this foundation, Li et al. introduce two large-scale benchmarks, RGBT234\cite{li2019rgb} and LasHeR~\cite{li2021lasher}, which significantly promote the development of RGBT tracking by providing high-quality, diverse annotations.
In object detection, researchers design robust RGBT-based methods for person~\cite{hwang2015multispectral, zhang2023drone, zhang2025transformer} and vehicle~\cite{sun2022drone, yuan2022translation} detection. Meanwhile, the first RGBT tiny object detection benchmark, RGBT-Tiny~\cite{ying2024visible}, focuses on the challenges of detecting small-scale objects in low-visibility conditions.
In the segmentation domain, Ji et al.\cite{ji2023multispectral} establish the large-scale MVSeg benchmark for RGBT video semantic segmentation and design an efficient fusion module that improves the effectiveness of cross-modal feature integration, thereby enabling more robust semantic understanding.
Additionally, RGBT fusion proves beneficial in saliency detection and person re-identification tasks, as demonstrated by a series of studies\cite{tu2022rgbt, tang2019rgbt, zheng2021robust, wang2022interact}, which show that thermal information enhances discriminative capability in complex scenes.
Overall, the thermal modality consistently strengthens the robustness and generalization ability of visual models across various domains, especially in challenging environments.
However, despite these advancements, most existing efforts overlook the potential of thermal information in cross-modal retrieval tasks involving textual descriptions.
To address this gap, we propose the Text-RGBT person retrieval task, which leverages thermal signals to enable accurate and robust person retrieval under arbitrary illumination conditions.

\section{Method}
\begin{figure*}[htbp]
\centering\includegraphics[width=6.9in,height=2.8in]{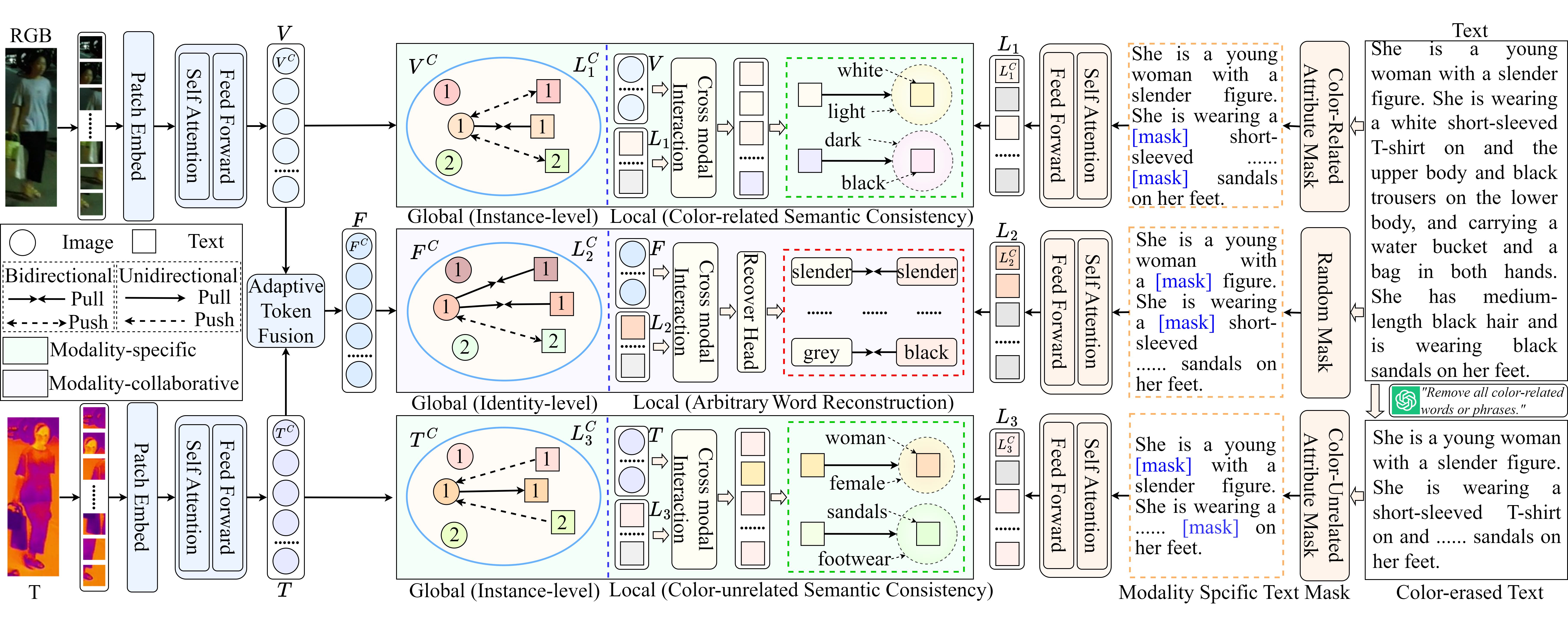}
\caption{DCAlign framework overview. DCAlign takes paired RGB and T images along with a complete text as input. On the visual side, modality-specific features are extracted using a shared encoder and fused through an adaptive token fusion module to generate joint representations. On the textual side, the complete text is processed with modality-specific masking strategies to produce three types of descriptive captions, which are then encoded into corresponding embeddings. The framework jointly models modality-collaborative and modality-specific visual-text alignment, optimizing the semantic alignment between RGBT multi-modal representations and text at both global and local levels.
}
\label{fig_2}
\end{figure*}
\subsection{Overview}
Figure~       \ref{fig_2} shows the architecture of DCAlign, a framework that takes paired RGBT person images and their corresponding text captions as input.
First, the RGB and T images are divided into non-overlapping patches, and their feature embeddings $V$ and $T$ are extracted using a shared image encoder.
To obtain richer multi-modal visual representations, we introduce an adaptive token fusion module to fuse the embeddings from both modalities, producing a fused RGBT representation $F$.
For the text branch, we first utilize ChatGPT-3.5 to remove all color-related words or phrases from the original caption, producing a color-erased version.
Subsequently, both the original and color-erased captions are processed with modality-specific text masking strategies to generate three distinct text variants: a color-related masked text, a random masked text, and a color-unrelated masked text.
These are then encoded using a shared text encoder, resulting in three textual feature embeddings: $L_1$, $L_2$, and $L_3$.
Finally, the three visual features $V$, $F$, and $T$ and the corresponding textual features $L_1$, $L_2$, and $L_3$ are fed into the proposed decoupled cross-modal alignment module, where both modality-collaborative and modality-specific visual-text alignment are performed to enhance semantic correspondence across modalities.
\subsection{Semantic Attribute Layering and Masking}
To fully exploit the complementary strengths of RGB and thermal modalities in attribute perception, we introduce a modality-specific text masking mechanism to facilitate hierarchical structuring and selective masking of semantic attributes in the text.
Specifically, we categorize textual semantics into three levels based on their dependency on color information: color-related attributes, color-unrelated attributes, and arbitrary attributes.
For these different semantic levels, we design three corresponding masking strategies to generate structured semantic variants that guide the model’s alignment behavior. 
The Color-Related Masking (CRM) randomly masks only the color-descriptive words in the text while keeping other semantics intact, encouraging the model to focus on aligning with RGB image features that are sensitive to color. 
The Color-Unrelated Masking (CUM) first uses a language model (e.g., ChatGPT) to automatically remove all color expressions from the text, then applies random masking to the nouns in the remaining text, guiding the model to focus on shape, structure, and appearance attributes better captured by the thermal modality. 
The Random Masking (RM) randomly masks words throughout the entire text without distinguishing attribute categories, encouraging the model to comprehensively utilize the complementary characteristics of RGB and thermal modalities for more robust semantic alignment.
Although this masking mechanism does not directly enforce semantic disentanglement, the modality-specific perturbations introduced during training enable the model to implicitly learn to differentiate and attend to various attribute types. 
This mechanism lays a solid foundation for the subsequent fine-grained modality-specific and modality-collaborative alignment strategies.

\subsection{Decoupled Cross-Modal Alignment}
Building on the semantic layering and masking from the previous section, we propose a decoupled cross-modal alignment strategy. We use the fused RGBT representation for robust alignment of arbitrary semantics, while designing modality-specific strategies for RGB and T to handle color-related and color-unrelated attributes, enabling fine-grained cross-modal coordination.
To this end, we decompose the alignment strategy into two branches: modality-collaborative alignment for arbitrary semantics, and modality-specific alignment for color-sensitive or invariant attributes, as detailed below.

\subsubsection{Modality-collaborative Visual-Text Alignment}
\noindent
At this semantic level, we aim to enhance the model's capacity to represent and understand arbitrary attributes that are not constrained by structural patterns or color dependencies. 
To this end, we introduce a dual alignment strategy that combines identity-level global alignment and token-level local reconstruction, enabling robust and fine-grained cross-modal semantic learning.

\paragraph{\textbf{Global Alignment}: Similarity Distribution Matching (SDM)}

\noindent
To align visual and textual representations at the identity level, we adopt a Similarity Distribution Matching (SDM) strategy. 
Let $F$ denote the fused RGBT multi-modal embedding, and $F^C$ its corresponding global feature. 
The masked textual input $T_2$ is encoded into a global representation $L_2^C$. 
We compute pairwise cosine similarities between $F^C$ and $L_2^C$ and apply softmax normalization to obtain the fusion-to-text similarity distribution $p_{i,j}$:

\begin{equation}
\label{deqn_ex1}
\begin{split}
p_{i,j} = \frac{\exp(sim(F_{i}^C, {L_j}_2^C) / \tau_1)}{\sum\nolimits_{k=1}^{N} \exp(sim(F_{i}^C, {L_k}_2^C) / \tau_1)},
\end{split}
\end{equation}

\noindent
where $sim(\cdot,\cdot)$ denotes cosine similarity, $\tau_1$ is a temperature parameter controlling the distribution sharpness, and $N$ is the batch size. 
To enforce distribution-level alignment, we compute the Kullback-Leibler (KL) divergence between the predicted and ground-truth similarity distributions $q_{i,j}$:

\begin{equation}
\label{deqn_ex2}
\begin{split}
L_{f2t} = \frac{1}{N}\sum\nolimits_{i=1}^{N}\sum\nolimits_{j=1}^{N} p_{i,j} \log \frac{p_{i,j}}{q_{i,j} + \epsilon},
\end{split}
\end{equation}

\noindent
where $\epsilon$ is a small constant for numerical stability. Similarly, we compute the reverse text-to-fusion loss $L_{t2f}$, and define the final SDM loss as:

\begin{equation}
\label{deqn_ex3}
\begin{split}
L_{SDM} = L_{f2t} + L_{t2f}.
\end{split}
\end{equation}

\paragraph{\textbf{Local Alignment}: Arbitrary Word Reconstruction (AR)}

\noindent
To further capture fine-grained semantics, we introduce a token-level local alignment strategy via arbitrary word reconstruction. 
Given the fused visual representation $F$ and the masked text $L_2$, we pass both through a Cross-modal Interaction Module, which integrates multi-head cross attention (with $L_2$ as query, $F$ as key and value) followed by two stacked Transformer blocks. 
This yields a contextualized output $\{{R_2}_i\}_{i=1}^{l}$, where $l$ is the sequence length.
A Recover Head maps each token to the vocabulary space. 
We compute cross-entropy loss over masked token positions $M_2$, using ground-truth tokens $G_{2i}$ to supervise reconstruction:

\begin{equation}
\label{deqn_ex8}
\begin{split}
L_{AR} = \sum\limits_{i\in M_2} \mathrm{CE}(\mathrm{MLP}(R_{2i}), G_{2i}),
\end{split}
\end{equation}

\noindent
where $\mathrm{MLP}$ denotes the Recover head, and $\mathrm{CE}$ is the cross-entropy loss. 

\subsubsection{Modality-specific Visual-Text Alignment}
The RGB and T modalities capture person-related information from different perspectives, providing distinct contributions to text  alignment. 
Accordingly, we design separate alignment mechanisms for text with the RGB modality and for color-erased text with the T modality.
Each mechanism includes both global and local alignment components to achieve fine-grained and robust cross-modal semantic matching.

\paragraph{\textbf{Global Alignment}: Bidirectional Instance Alignment (BIA) and Unidirectional Instance Binding (UIB).}
The BIA and UIB modules are designed to model the global semantic alignment between RGB images and text, and between T images and color-erased text, respectively.
Given the inherent limitations of a single visual modality in representing complete semantics, enforcing identity-level cross-modal alignment may lead to unstable feature spaces, ambiguous semantic correspondence, or even optimization difficulty.
To address this, we adopt instance-level global alignment strategies in both semantic levels to preserve the stability of the shared embedding space while enabling effective semantic transfer.
Specifically, modern vision-language pretraining models (e.g., CLIP) are trained on large-scale RGB-text pairs and exhibit strong semantic alignment capabilities. 
To leverage this prior knowledge, BIA introduces a bidirectional InfoNCE loss to align the RGB representation $V$ with text features $L_1$, encouraging mutual semantic consistency. 
The loss functions as follows:
\begin{equation}
\begin{aligned}
L_{V,L_1} &= -\frac{1}{N} \sum_{i=1}^{N} \log \frac{\exp\left(\text{sim}(V_i^{C}, L_{i1}^{C}) / \tau_2\right)}{\sum_{j=1}^{N} \exp\left(\text{sim}(V_i^{C}, L_{j1}^{C}) / \tau_2\right)}, \\
L_{L_1,V} &= -\frac{1}{N} \sum_{i=1}^{N} \log \frac{\exp\left(\text{sim}(L_{i1}^{C}, V_i^{C}) / \tau_2\right)}{\sum_{j=1}^{N} \exp\left(\text{sim}(L_{i1}^{C}, V_j^{C}) / \tau_2\right)}, \\
L_{\text{BIA}} &= L_{V,L_1} + L_{L_1,V}.
\end{aligned}
\end{equation}
Here, $ V_i^{C} $ denotes the global embedding of the $i$-th RGB image, $ L_{i1}^{C} $ is the global embedding of the text, $ \text{sim}(\cdot, \cdot) $ denotes cosine similarity, $ \tau_2 $ is a temperature coefficient, and $ N $ is the batch size.

In UIB, considering the thermal modality’s sensitivity to structural and contour features, we introduce a unidirectional binding strategy inspired by ImageBind. 
This strategy pulls the thermal embedding $ T_i^{C} $ toward a fixed embedding space $ L_{i3}^{C} $, which corresponds to the global text representation with color-related words removed. 
The loss is defined as:
\begin{equation}
L_{\text{UIB}} = -\frac{1}{N} \sum_{i=1}^{N} \log \frac{\exp\left(\text{sim}(T_i^{C}, L_{i3}^{C}) / \tau_2\right)}{\sum_{j=1}^{N} \exp\left(\text{sim}(T_i^{C}, L_{j3}^{C}) / \tau_2\right)}.
\end{equation}
Here, $ L_{i3}^{C} $ is the global text embedding with color-related words removed. To preserve the pretrained semantic space, $ L_{i3}^{C} $  is frozen during training.

\paragraph{\textbf{Local Alignment}: Color-Related Semantic Consistency (CRS) and Color-Unrelated Semantic Consistency (CUS).}
CRS and CUS are designed to achieve fine-grained semantic alignment between the RGB modality and color-related attributes, and between the T modality and color-independent attributes, respectively. 
However, due to the inherent limitations of a single visual modality in semantic representation, applying strict matching constraints on the reconstruction results may lead to inadequate semantic fitting, which limits the model’s ability to capture attribute details and hinders the modeling of complex semantic structures.  
To address this issue, we introduce a semantic consistency loss, which enables a soft alignment strategy. 
This approach preserves the stability of global semantics while enhancing the model’s capacity for fine-grained attribute modeling and alignment.

Specifically, in the CRS branch, we feed the RGB embedding \( V \) and the color-related text \( L_1 \) into a cross-modal interaction module to obtain the reconstructed representation of each token, denoted as \( \{ R_{1i} \}_{i=1}^l \). 
We then compute the cosine similarity between these reconstructed tokens and their corresponding ground truth semantic embeddings \( \{ E_{1i} \}_{i=1}^l \) (i.e., the true token embeddings from the complete, unmasked text) at the masked positions \( M_1 \), and define the semantic consistency loss as:
\begin{equation}
\label{deqn_ex11}
L_{CRS} = 1 - \text{Avg}\left( \sum_{i \in M_1} \text{sim}(R_{1i}, E_{1i}) \right).
\end{equation}
This loss encourages the reconstructed tokens to be semantically aligned with the ground truth, leading to a more robust and semantically-aware reconstruction process.
The CUS branch shares a similar structure with CRS. The difference lies in that it takes the thermal embedding \( T \) and the color-unrelated text \( L_3 \) as input to the cross-modal interaction module, generating the reconstructed token representations \( \{ R_{3i} \}_{i=1}^l \). 
We then compute their similarity to the corresponding ground truth embeddings \( \{ E_{3i} \}_{i=1}^l \) (from the complete version of \( L_3 \)) at the masked positions \( M_3 \), and define the loss as:
\begin{equation}
\label{deqn_ex12}
L_{CUS} = 1 - \text{Avg}\left( \sum_{i \in M_3} \text{sim}(R_{3i}, E_{3i}) \right).
\end{equation}
This loss guides the thermal modality to focus on the semantic structure of color-independent attributes, thereby enhancing the fine-grained alignment between RGBT multi-modal representations and text.

\subsection{Adaptive Token Fusion}
\begin{figure}[!t]
\centering
\includegraphics[width=3.1in,height=1.6in]{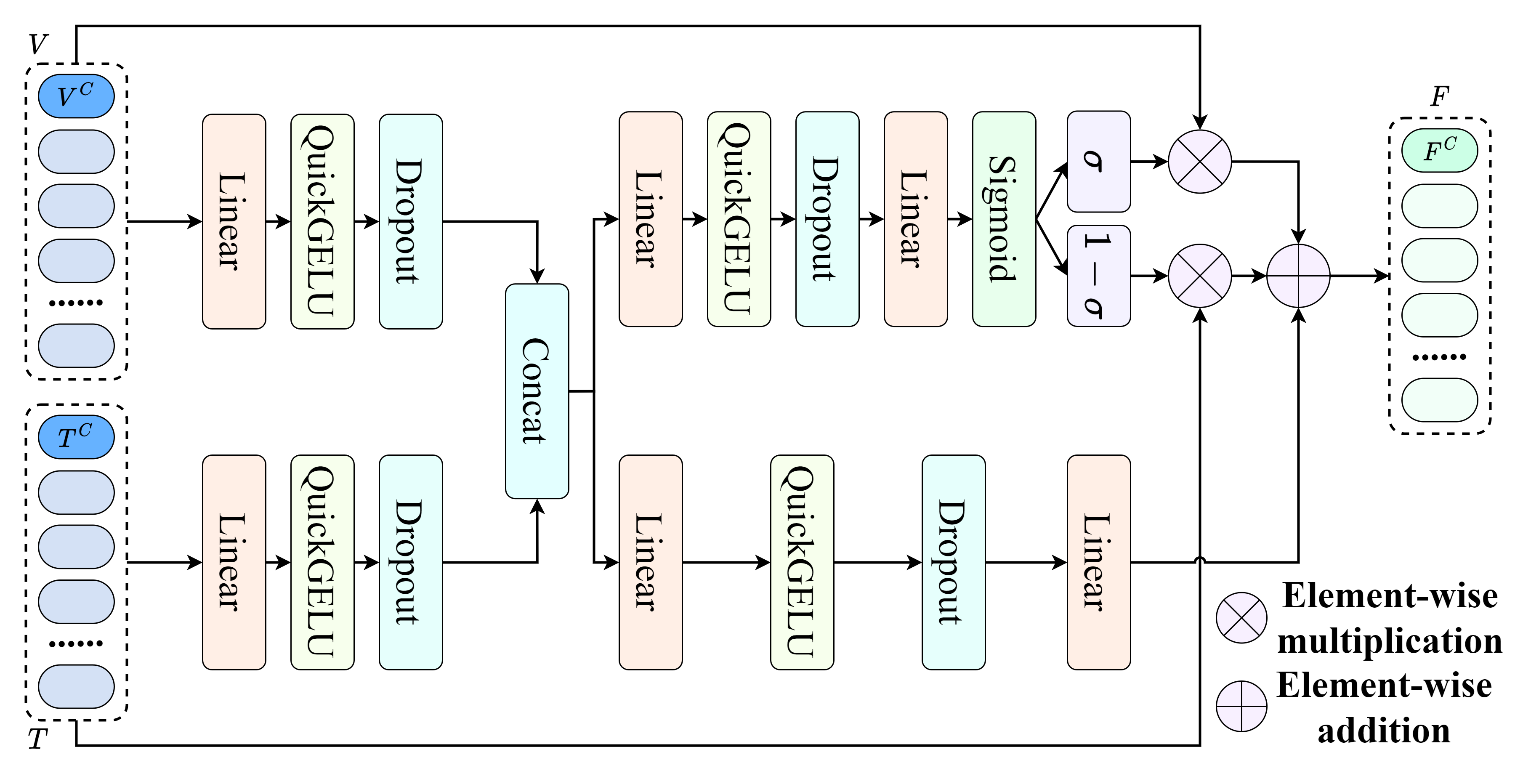}
\caption{Overview of the Adaptive Token Fusion module. It takes visual embeddings $V$ and $T$ as input and generates a fused visual embedding $F$ via an adaptive fuse mechanism.}
\label{fig_3}
\end{figure}
As illustrated in Figure~\ref{fig_3}, the proposed Adaptive Token Fusion (ATF) module dynamically integrates modality-specific information from the visible and thermal streams. 
It takes as input the embeddings $V$ and $T$ from the visual encoder, which are first passed through two linear layers with QuickGELU and Dropout to enhance feature expressiveness.
The transformed features are concatenated into a joint RGBT embedding, capturing complementary information from both modalities.
This embedding is then fed into two parallel branches with non-shared parameters.
The first branch predicts adaptive fusion weights via a Sigmoid-activated network to guide a convex combination of the original RGB and T embeddings, enabling the model to dynamically emphasize the more informative modality.
The second branch refines the joint embedding to enhance cross-modal interaction and semantic alignment.
Finally, the fused RGBT embedding and the refined joint embedding are combined via element-wise addition to produce a robust, context-aware representation. 
This design allows ATF to adaptively balance modality contributions and improves alignment between text and visual features under complex conditions.

\section{Benchmark}
\subsection{Data Collection and Annotations}
\subsubsection{\textbf{Data Capture and Processing}}
We employ the DJI Matrice 350 RTK drone as our data acquisition platform, equipped with the Zenmuse H20T multi-sensor gimbal camera. 
This setup enables the simultaneous capture of high-quality visible and thermal imagery, with RGB images at a resolution of 1080×1920 and thermal images at 512×640.
To simulate realistic surveillance scenarios, the drone operates at an altitude of 4 to 8 meters—consistent with common urban roadside monitoring setups—ensuring that the collected pedestrian images reflect practical deployment conditions.
For the collected RGBT video sequences, we first filter out those with errors or extremely short durations.
We then manually align the RGB and thermal videos using Adobe Photoshop to achieve spatial registration between modalities. 
From the registered sequences, frames are sampled at a rate of 2–3 frames per second, resulting in paired RGBT images at a resolution of 640×512.
Finally, we utilize YOLO v8 to detect pedestrians in the thermal images. 
The detected bounding boxes are then used to crop the corresponding regions in the RGB images, forming aligned RGBT person image pairs. 
By combining both modalities for identity classification, we construct a multimodal pedestrian dataset containing 1,822 identities and 4,723 image pairs.
As shown in Figure~\ref{fig_10}, several examples of paired RGBT images of different pedestrians are presented, intuitively reflecting the multimodal characteristics of the dataset.

\subsubsection{\textbf{Caption Annotations}}
\begin{figure}[htbp]
\centering
\includegraphics[width=3.1in,height=3.8in]{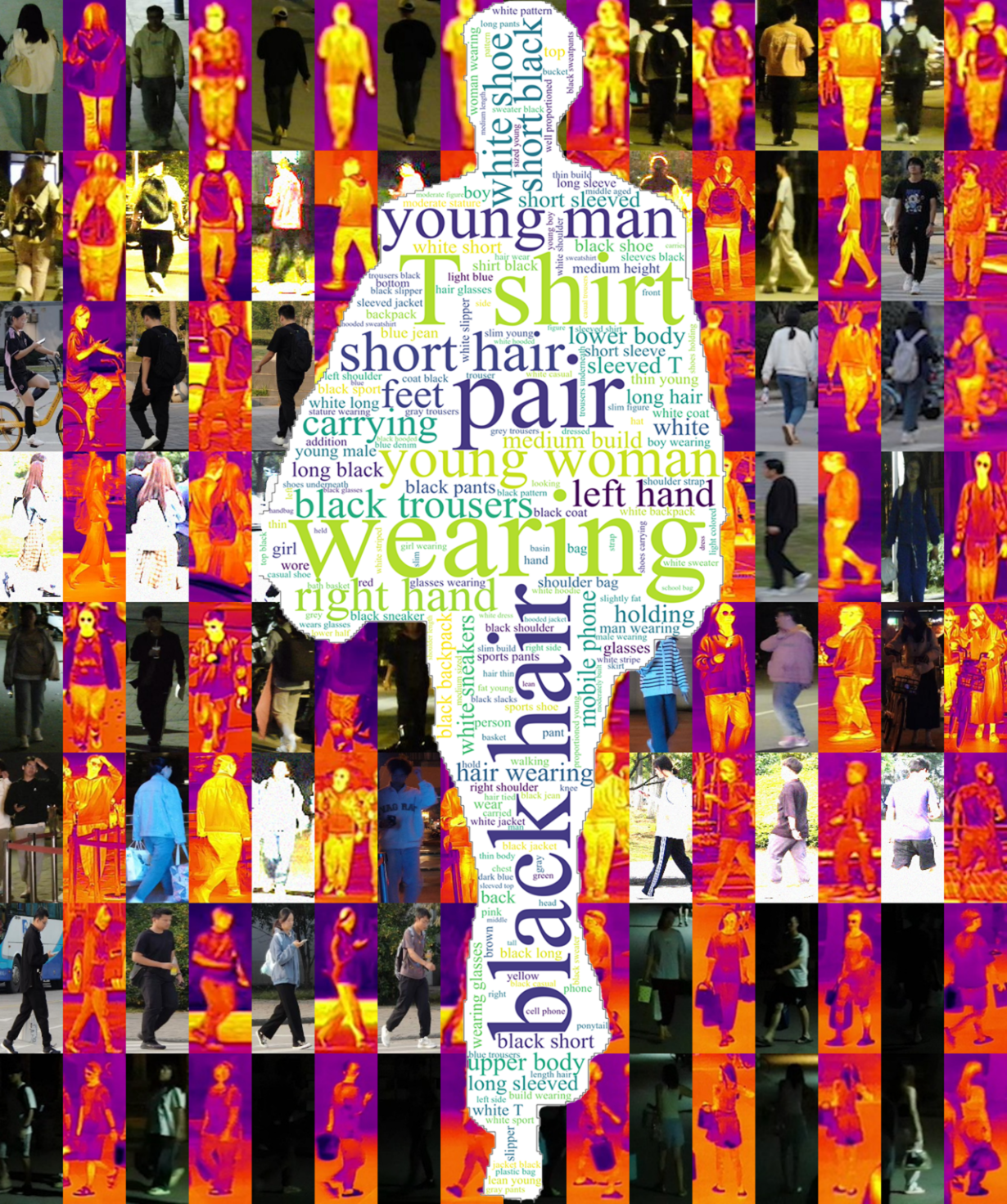}
\caption{Paired RGBT person images and high-frequency words in our dataset.}
\label{fig_10}
\end{figure}

\begin{figure}[htbp]
\centering
\includegraphics[width=3.1in,height=2in]{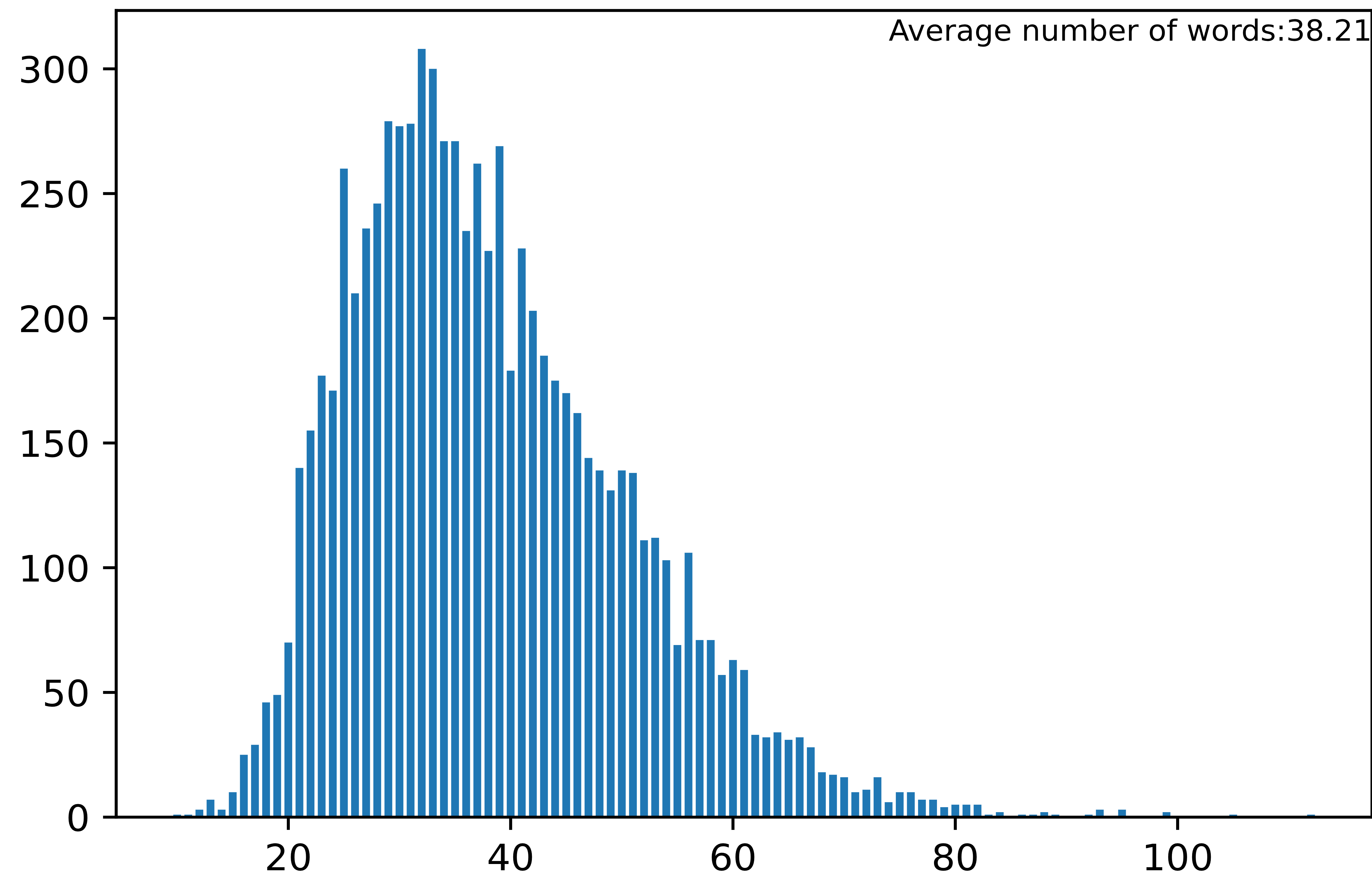}
\caption{Distribution Statistics of Word Counts in Text}
\label{fig_11}
\end{figure}

To ensure accurate and diverse textual descriptions, we train a team of 23 annotators to carefully describe person images.
Given the unique characteristics of our dataset, we could not directly adopt annotation strategies used in conventional text-image person retrieval datasets, which typically consist of well-lit RGB images. 
In such datasets, person attributes can be easily identified from the RGB modality alone. 
However, our benchmark includes challenging scenarios such as low-light conditions, where RGB imagery alone may fail to capture critical appearance details.
To overcome this limitation, we leverage both RGB and T modalities to provide richer and more complete annotations. 
Annotators are instructed to focus on attributes such as gender, body shape, age, hairstyle, clothing (type and color), carried objects, and accessories.
Each annotator is assigned a unique set of images to encourage personalized and varied descriptions. 
Moreover, to accommodate real-world application scenarios, we vary the number of descriptions per image: 3,264 images were annotated with two descriptions, while 1,459 had one.
After several rounds of manual review and validation against the aligned RGBT image pairs, we obtain a total of 7,987 high-quality textual descriptions, each accurately reflecting the visual content and fine-grained attributes of the corresponding person image.
The distribution of high-frequency textual words in the entire dataset is shown in Figure~\ref{fig_10}, intuitively illustrating the common key attributes and descriptive terms. 
Additionally, Figure~\ref{fig_11} presents the statistical distribution of the number of words per text description, reflecting the length and level of detail in the annotations, which further demonstrates the richness and diversity of our dataset.

\subsubsection{\textbf{Training and Test Sets}}
We split the dataset into training and testing sets following a 7:3 ratio, resulting in a balanced and representative partition.
Instead of performing a purely random split, we adopt a challenge-oriented partitioning strategy, where the data is divided within each predefined challenge category.
This ensures that all challenge types are proportionally represented in both training and testing sets, enabling the model to learn robust representations while also being evaluated under diverse and realistic conditions.
After the split, the training set consists of 1,266 distinct identities, 3,250 RGBT person image pairs, and 5,497 corresponding textual descriptions.
The testing set comprises 553 identities, 1,473 image pairs, and 2,490 textual descriptions, providing a comprehensive basis for evaluating the generalization ability of models across multiple scene challenges.
\begin{figure*}[htbp]
\centering
\includegraphics[width=6.8in,height=1.7in]{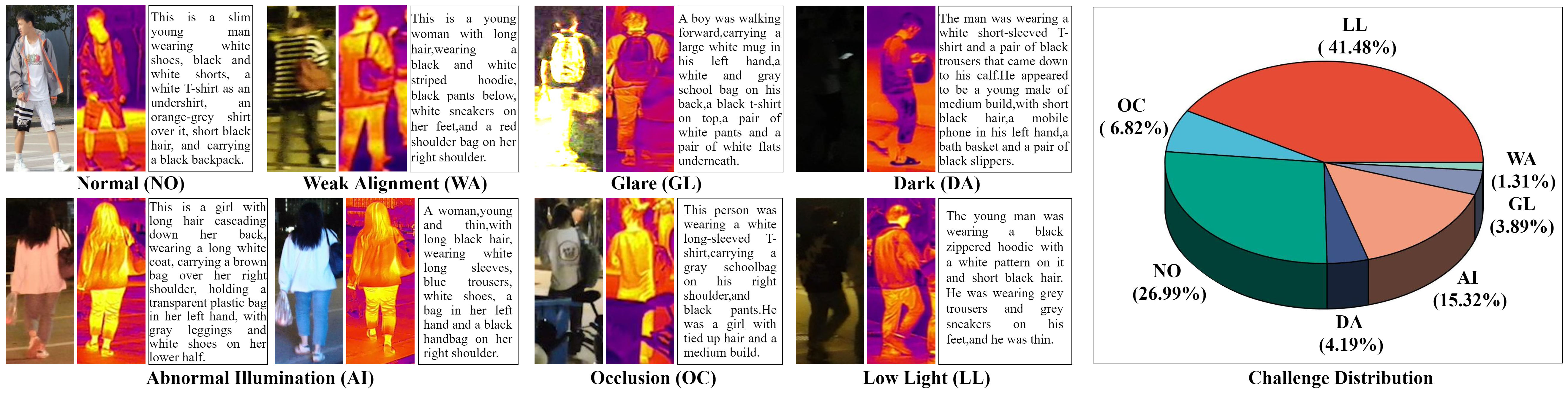}
\caption{Demonstration of different challenging scenarios in RGBT-PEDES.}
\label{fig_5}
\end{figure*}
\subsection{Benchmark Properties and Statistics}
RGBT-PEDES is the first benchmark dataset to introduce the thermal modality into the text-image person retrieval task, addressing the limitations of existing datasets that rely solely on visible images.
It comprises 1,822 unique identities, with each identity associated with multiple paired RGB and thermal images captured from diverse camera views and under varying environmental conditions, such as low illumination, occlusion, and background clutter.
Each image pair is annotated with 1–2 natural language descriptions that depict fine-grained details, including appearance attributes and carried objects, thus enabling fine-level alignment between textual and visual modalities.
To provide a comprehensive understanding of the dataset, we introduce RGBT-PEDES from two complementary perspectives:
(1) Image characteristics, highlighting the diversity of viewpoints, lighting conditions, and the complementary nature of RGB and thermal modalities. 
(2) Caption characteristics, focusing on the descriptive richness, sentence length distribution, and the presence of fine-grained semantic attributes that are crucial for retrieval.
Overall, RGBT-PEDES establishes a new and challenging benchmark for advancing research in cross-modal person retrieval, particularly in complex real-world scenarios.
\begin{table}[htbp]
\caption{Comparison RGBT-PEDES with other datasets for text-image person retrieval.}
\renewcommand{\arraystretch}{1.3}
\setlength{\tabcolsep}{0.15cm}{
\begin{tabular}{c|cc|cc}
\hline
\multirow{2}{*}{Dataset} & \multicolumn{2}{c|}{Image}   & \multicolumn{2}{c}{Caption}           \\ \cline{2-5} 
                         & \multicolumn{1}{c|}{RGB} & T & \multicolumn{1}{c|}{Maximum} & Average \\ \hline
CUHK-PEDES               & \multicolumn{1}{c|}{40,206}    & 0   & \multicolumn{1}{c|}{96}      & 23.5    \\ 
ICFG-PEDES               & \multicolumn{1}{c|}{54,522}    & 0  & \multicolumn{1}{c|}{83}      & 37.2    \\ 
RSTPReid                 & \multicolumn{1}{c|}{20505}    & 0  & \multicolumn{1}{c|}{70}      & 26.5    \\ \hline
RGBT-PEDES               & \multicolumn{1}{c|}{4723}    & 4723  & \multicolumn{1}{c|}{112}     & 38.2    \\ \hline
\end{tabular}
}
\label{tab_1}
\end{table}
\subsubsection{Visual and Caption Characteristics}
As shown in Table~\ref{tab_1}, RGBT-PEDES is the first dataset to introduce the thermal modality in text-image person retrieval.  
Collected from diverse scenes with complex lighting, this dataset presents greater challenges for cross-modal retrieval. 
To thoroughly analyze it, we categorized all image pairs based on different challenges.  
RGBT-PEDES includes not only common scenes but also special cases such as weak alignment causing feature matching difficulties, glare affecting visibility, lack of visible light in dark environments, image quality instability due to abnormal lighting, occlusions, and degraded visible image quality under low light.
These scenarios are shown in Fig.~\ref{fig_5}.  

RGBT-PEDES contains 7,987 textual descriptions, with 3,264 images having two descriptions each and 1,459 having one. This design better reflects real-world variations in text queries.  
Compared to other text-image person retrieval datasets, our captions are more detailed, with the longest reaching 112 words and the highest average word count per sentence.

\begin{table*}[htbp]
\renewcommand{\arraystretch}{1.3}
\setlength{\tabcolsep}{0.19cm}{
\caption{Performance comparison with state-of-the-art methods on the RGBT-PEDES dataset. "G+L" in the "Type" column indicates that the method incorporates both global and local alignment strategies. In the "Fusion" column, "Add" denotes feature fusion by element-wise addition of RGB and T image features, while "Cat" denotes feature fusion by concatenation.~\label{tab:table3}}
\begin{tabular}{c|cccc|cccccc}
\hline
\multicolumn{1}{c|}{Method}                 & \multicolumn{1}{c|}{Type} & \multicolumn{1}{c|}{Ref} & \multicolumn{1}{c|}{Image Enc.} & Text Enc.             & Fusion & Rank-1 & Rank-5 & Rank-10 & mAP   & RSum   \\ \hline
\multicolumn{1}{c|}{\multirow{2}{*}{NAFS~\cite{gao2021contextual}}}  & \multirow{2}{*}{G+L}        & \multirow{2}{*}{arXiv21} & \multirow{2}{*}{ResNet50}           & \multirow{2}{*}{BERT} & Add    & 25.83  & 50.31  & 60.99   & -     & 137.13 \\ 
\multicolumn{1}{c|}{}                       &                           &                          &                                 &                       & Cat    & 26.51  & 50.41  & 60.35   & -     & 137.27 \\ \hline
\multicolumn{1}{c|}{\multirow{2}{*}{TIPCB~\cite{chen2022tipcb}}} & \multirow{2}{*}{G+L}        & \multirow{2}{*}{Neuro22} & \multirow{2}{*}{ResNet50}           & \multirow{2}{*}{BERT} & Add    & 16.11  & 36.94  & 49.07   & 15.03 & 102.12 \\
\multicolumn{1}{c|}{}                       &                           &                          &                                 &                       & Cat    & 15.74  & 34.98  & 47.55   & 14.96 & 98.27  \\  \hline
\multirow{2}{*}{LGUR~\cite{shao2022learning}}                        & \multirow{2}{*}{G+L}        & \multirow{2}{*}{MM22}    & \multirow{2}{*}{DeiT-Small}     & \multirow{2}{*}{BERT} & Add    & 27.63  & 52.97  & 64.62   & 24.49 & 145.22 \\
                                             &                           &                          &                                 &                       & Cat    & 22.09  & 46.59  & 59.48   & 19.73 & 128.16 \\ \hline
\multirow{2}{*}{APTM~\cite{yang2023towards}}                        & \multirow{2}{*}{G+L}        & \multirow{2}{*}{MM23}    & \multirow{2}{*}{Swin-B}     & \multirow{2}{*}{BERT} & Add    &    53.15    &     78.54    &    85.94     &  49.62     &  217.63      \\
                                             &                           &                          &                                 &                       & Cat    &   50.71     & 75.91       & 83.29  &  46.76     &   209.91     \\ \hline
\multirow{2}{*}{IRRA~\cite{jiang2023cross}}                        & \multirow{2}{*}{G+L}        & \multirow{2}{*}{CVPR23}    & \multirow{2}{*}{CLIP-ViT}     & \multirow{2}{*}{CLIP-Xformer} & Add    &        55.62 & 80.14 &  87.91 & 52.43 & 223.67    \\ 
                                             &                           &                          &                                 &                       & Cat    &   47.72 & 74.26 & 83.95 & 46.23 & 205.93       \\ \hline
\multirow{2}{*}{PLIP~\cite{zuo2024plip}}                        & \multirow{2}{*}{G+L}        & \multirow{2}{*}{NIPS24}    & \multirow{2}{*}{ResNet50}     & \multirow{2}{*}{BERT} & Add    & 26.41 &    48.74    &   59.44      & 23.04       &  134.59       \\ 
                                             &                           &                          &                                 &                       & Cat    &   21.39     &   44.13     &  55.57       &  19.94     &   120.89     \\ \hline
\multirow{2}{*}{CFAM~\cite{zuo2024ufinebench}}                        & \multirow{2}{*}{G+L}        & \multirow{2}{*}{CVPR24}    & \multirow{2}{*}{CLIP-ViT}     & \multirow{2}{*}{CLIP-Xformer} & Add    &  55.46 & 80.60 & 88.15 & 52.76 & 224.21     \\ 
                                             &                           &                          &                                 &                       & Cat    & 50.12 & 77.23 & 86.43 & 48.61 & 213.78    \\ \hline
\multirow{2}{*}{TBPS-CLIP~\cite{cao2024empirical}}                        & \multirow{2}{*}{G+L}        & \multirow{2}{*}{AAAI24}    & \multirow{2}{*}{CLIP-ViT}     & \multirow{2}{*}{BERT} & Add    &  51.81 & 71.81 & 79.75 & 48.61 & 203.37     \\ 
                                             &                           &                          &                                 &                       & Cat    & 29.47 & 48.59 & 59.59 & 30.39 & 137.65       \\ \hline
Baseline                                         & G                         & -                        & CLIP-ViT                        & CLIP-Xformer          & Add &  54.90 & 80.00 & 88.19 & 52.19 & 223.09 \\ \hline
Ours                                         & G+L                         & -                        & CLIP-ViT                        & CLIP-Xformer          & -      &  \red{58.47} & \red{82.01} & \red{89.52} & \red{54.59} & \red{230.00}       \\ \hline
\end{tabular}
}
\end{table*}

\section{Experiments}
In this section, we extend seven text-image person retrieval methods (NAFS~\cite{gao2021contextual}, TIPCB~\cite{chen2022tipcb}, LGUR~\cite{shao2022learning}, APTM~\cite{yang2023towards}, IRRA~\cite{jiang2023cross}, PLIP~\cite{zuo2024plip}, CFAM~\cite{zuo2024ufinebench} and TBPS-CLIP~\cite{cao2024empirical}) to the Text-RGBT task, and conduct experiments on the proposed RGBT-PEDES dataset to evaluate the overall effectiveness of DCAlign and the contribution of each loss function.
\subsection{Evaluation Metric}
\textbf{Rank-k} is a metric used to evaluate the retrieval performance by determining whether the correct person is ranked within the top k positions in the retrieval results. Rank-1 measures the accuracy of the retrieval system by checking if the correct person is placed in the first position. If the correct person is ranked first, the score is 1; otherwise, it is 0. The final Rank-1 value is obtained by averaging the results across all queries, providing a clear indication of how often the correct person is retrieved as the top result.
Similarly, Rank-5 evaluates the system's ability to retrieve the correct person within the top 5 positions of the search results, while Rank-10 checks if the correct person appears within the top 10 positions. These metrics provide a broader view of the system's performance, particularly in scenarios where a perfect match is not expected to always rank first but may appear within a set of relevant results. The Rank-5 and Rank-10 values are also averaged across all queries, reflecting the overall efficiency of the retrieval system in returning relevant results within the top positions.

\textbf{mAP} (mean Average Precision) is a comprehensive metric for evaluating the performance of a retrieval system, calculating the average precision across all queries while considering ranking. It assesses both the ability to retrieve the correct target and the matching precision across multiple results. Unlike Rank-based metrics, mAP reflects performance at different recall rates, balancing precision and recall.
Specifically, mAP first calculates the Average Precision (AP) for each query, where AP is computed based on the ranking of retrieval results, considering the accuracy of each result's position. Then, the AP values of all queries are averaged to obtain the final mAP score. mAP not only evaluates the system's ability to return relevant results in the top positions but also focuses on maintaining high retrieval precision across the entire retrieval process.

\textbf{RSum} is the sum of Rank-1, Rank-5, and Rank-10. It evaluates the effectiveness of a retrieval system by considering the rankings at three different levels: whether the correct result is ranked first (Rank-1), within the top 5 results (Rank-5), and within the top 10 results (Rank-10). By combining these three rank-based metrics, RSum provides a more comprehensive view of the system's retrieval accuracy, with higher values indicating better overall performance in returning relevant results across different ranks.

\subsection{Implementation Details}
DCAlign leverages the image and text encoders of CLIP to extract features from person images and captions. Both the RGB and T modalities share weights in the image encoders, while the text encoder extracts three types of masked text features. 
During training, the model applies standard data augmentation techniques: random horizontal flipping, random cropping with padding, and random erasing to improve robustness. Input images are resized to $384 \times 128$ pixels and divided into 16×16 patches. 
The hidden layer dimension is set to 512, and the text descriptions are serialized with a maximum length of 77 tokens.
In line with previous works~\cite{radford2021learning}~\cite{jiang2023cross}, the hyperparameters $\tau_1$ and $\tau_2$ in the DCALign are set to 0.07 and 0.02, respectively. 
The training process consists of two stages. In Stage I, the visual modality tokens (from both RGB and T images) are fused using element-wise addition, with only the image and text encoders being updated during training. 
In Stage II, fusion is performed using adaptive token fusion, where all encoders are frozen, and only the fusion module is trained to adaptively combine the modality-specific features.
The Adam optimizer is employed, with an initial learning rate of $1 \times 10^{-5}$ for Stage I and $5 \times 10^{-5}$ for Stage II.
Both stages employ cosine annealing decay for the learning rate schedule to improve convergence. 
All experiments are conducted on a single RTX 4090 24GB GPU to ensure efficient training and evaluation.

\subsection{Comparison of Ours with Baseline}
\begin{figure*}[htbp]
\centering\includegraphics[width=6.6in,height=2.2in]{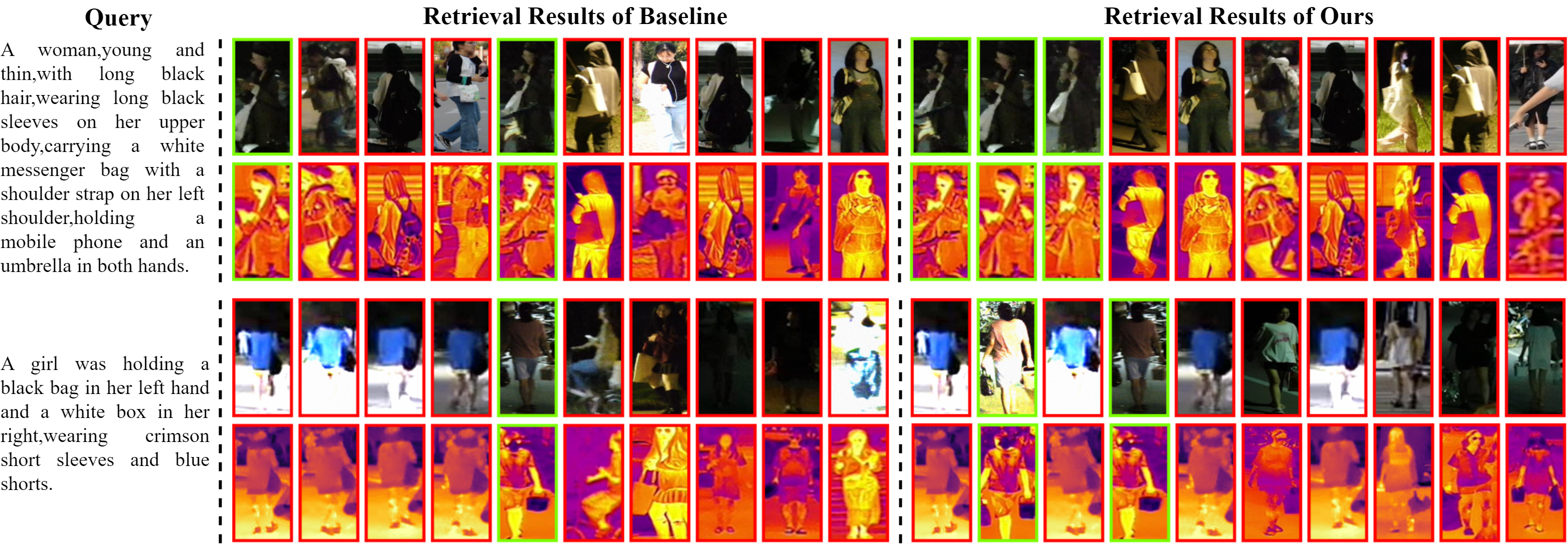}
\caption{Comparison of the top-10 retrieval results on RGBT-PEDES for each text query between the baseline and our method, with matching and mismatched images marked by green and red rectangles, respectively.}
\label{fig_6}
\end{figure*}
To verify the effectiveness of our proposed method, we compare it with a baseline method that employs a similarity distribution matching (SDM) loss between the text and RGBT fused features, where the fusion is performed via element-wise addition. 
As shown in Table~\ref{tab:table3}, the baseline achieves Rank-1, Rank-5, Rank-10, mAP, and RSum scores of 54.90\%, 80.00\%, 88.19\%, 52.19\%, and 223.09\%, respectively.
In contrast, our method achieves 58.47\%, 82.01\%, 89.52\%, 54.59\%, and 230.00\% on the same metrics, resulting in improvements of 6.5\% in Rank-1, 2.4\% in mAP, and 6.91\% in RSum, significantly outperforming the baseline across all evaluation metrics.
This comprehensive performance gain is attributed to our proposed multilevel multi-granularity cross-modal alignment framework and adaptive fusion mechanism.

Specifically, we construct modality-specific masked texts and perform alignment with RGB, T, and RGBT fused features at multiple granularities, capturing global semantic correspondence at a coarse level and enabling accurate matching of key attributes at a fine-grained level. 
This design allows the model to fully exploit the complementary strengths of different visual modalities: the RGB modality helps recover color-related attributes, the T modality provides attribute information independent of color, and the RGBT fused representation supports the recovery of arbitrary attributes.
In addition, the adaptive fusion module predicts a dynamic weight to perform content-adaptive fusion of RGB and T features. 
This enables the model to selectively attend to the most informative modality based on semantic context, thereby enhancing its ability to handle complex descriptions.
Qualitative results shown in Figure~\ref{fig_6} further validate the effectiveness of our method.
In several retrieval examples, the baseline fails to retrieve the correct image, whereas our method successfully identifies the correct target by precisely capturing the key semantics described in the text.
These results demonstrate that our method offers superior fine-grained semantic understanding and more robust cross-modal alignment.

\subsection{Comparison of Ours with Other Methods}
As shown in Table~\ref{tab:table3}, we conduct a comprehensive performance comparison between our proposed method and eight representative state-of-the-art Text-RGBT person retrieval approaches, including NAFS, TIPCB, LGUR, APTM, IRRA, PLIP, CFAM, and TBPS-CLIP.
On the Text-RGBT retrieval task, our method achieves Rank-1, mAP, and RSum scores of 58.47\%, 54.59\%, and 230.00\%, respectively, consistently outperforming all existing methods across all evaluation metrics and demonstrating superior overall performance and robust cross-modal retrieval capability.
Specifically, compared with IRRA, which adopts a modality-wise additive fusion strategy, our method improves Rank-1, mAP, and RSum by 2.85\%, 2.16\%, and 6.33\%, respectively; compared to APTM, which also relies on additive fusion, the gains further increase to 5.32\%, 4.97\%, and 20.09\%; relative to CFAM, our method achieves improvements of 5.42\%, 1.55\%, and 2.58\%. Moreover, compared with PLIP, we achieve an absolute improvement of 32.06\% on the Rank-1 metric, demonstrating substantial performance advantages.
These results clearly validate the effectiveness of our approach, especially when contrasted with traditional methods such as NAFS, TIPCB, and LGUR, which lack the benefits of large-scale multimodal pretraining and exhibit weaker semantic modeling and robustness.

We attribute the inferior performance of existing methods on the Text-RGBT task mainly to two factors.
First, most text-image person retrieval methods optimize feature representations solely within the visible spectrum, overlooking the complementary information provided by the thermal modality and thus struggling to bridge the semantic gap between visible and thermal domains. 
Second, these methods generally lack mechanisms to explicitly model modality discrepancies; in challenging conditions such as low light or strong heat, the modality of thermal contribution may even be treated as noise, weakening model discrimination.
In contrast, our method explicitly models modality-specific characteristics by aligning RGB features with text, T features with text, and the fused RGBT representation with text at multiple semantic levels.
By combining modality-specific and modality-collaborative global-local semantic alignment strategies, our framework effectively exploits the complementarity between RGB and thermal modalities. 
This leads to significantly improved semantic consistency and matching accuracy, enabling much more robust cross-modal retrieval performance in complex scenarios.

\subsection{Ablation Study}
We use SDM loss between the fused visual multi-modal embeddings and the randomly masked text embeddings as the baseline method.
\begin{table*}[]
\renewcommand{\arraystretch}{1.3}
\setlength{\tabcolsep}{0.25cm}{
\caption{ Ablation study of different modules on the RGBT-PEDES benchmark. No.0 denotes the baseline method.~\label{tab:table4}}
\begin{tabular}{c|cc|cc|cc|c|c|c|c|c|c}
\hline
\multirow{2}{*}{No.}                                   & \multicolumn{2}{c|}{Modality-Collaborative}                          & \multicolumn{4}{c|}{Modality-Specific}   & \multirow{2}{*}{ATF}  & \multirow{2}{*}{Rank-1} & \multirow{2}{*}{Rank-5} & \multirow{2}{*}{Rank-10} & \multirow{2}{*}{mAP} & \multirow{2}{*}{RSum}  \\ \cline{2-7}
                                                       & \multicolumn{1}{c|}{SDM} & \multicolumn{1}{c|}{AR} & \multicolumn{1}{c|}{BIA} & \multicolumn{1}{c|}{CRS} & \multicolumn{1}{c|}{UIB} & \multicolumn{1}{c|}{CUS} &                       &                         &                         &                          &                       \\ \hline
0 & \multicolumn{1}{c}{$\checkmark$}    &                         & \multicolumn{1}{c}{}    &                          & \multicolumn{1}{c}{}    &                          & \multicolumn{1}{c|}{} & \multicolumn{1}{c}{54.90}   & \multicolumn{1}{c}{80.00}   & \multicolumn{1}{c}{88.19}    & \multicolumn{1}{c}{52.19} & 223.09 \\ 
1 & \multicolumn{1}{c}{$\checkmark$}    &   $\checkmark$   & \multicolumn{1}{c}{}    &                          & \multicolumn{1}{c}{}    &                          & \multicolumn{1}{c|}{} & \multicolumn{1}{c}{55.30}   & \multicolumn{1}{c}{80.65}   & \multicolumn{1}{c}{88.83}    & \multicolumn{1}{c}{52.62} & 224.78   \\ \hline
2 & \multicolumn{1}{c}{$\checkmark$}    &   $\checkmark$   & \multicolumn{1}{c}{$\checkmark$}    &                          & \multicolumn{1}{c}{}    &                          & \multicolumn{1}{c|}{} & \multicolumn{1}{c}{56.66}   & \multicolumn{1}{c}{80.88}   & \multicolumn{1}{c}{88.15}    & \multicolumn{1}{c}{53.25} & 225.69 \\ 
3 & \multicolumn{1}{c}{$\checkmark$}    &   $\checkmark$   & \multicolumn{1}{c}{$\checkmark$}    &      $\checkmark$                     & \multicolumn{1}{c}{}    &                          & \multicolumn{1}{c|}{} & \multicolumn{1}{c}{57.23}   & \multicolumn{1}{c}{81.29}   & \multicolumn{1}{c}{88.51}    & \multicolumn{1}{c}{53.65} & 227.03 \\ 
4 & \multicolumn{1}{c}{$\checkmark$}    &   $\checkmark$   & \multicolumn{1}{c}{$\checkmark$}    &          $\checkmark$                & \multicolumn{1}{c}{$\checkmark$}    &                          & \multicolumn{1}{c|}{} & \multicolumn{1}{c}{57.99}   & \multicolumn{1}{c}{80.96}   & \multicolumn{1}{c}{88.35}    & \multicolumn{1}{c}{53.84} & 227.30 \\ 
5 & \multicolumn{1}{c}{$\checkmark$}    &   $\checkmark$   & \multicolumn{1}{c}{$\checkmark$}    &        $\checkmark$                  & \multicolumn{1}{c}{$\checkmark$}    &    $\checkmark$                      & \multicolumn{1}{c|}{} & \multicolumn{1}{c}{58.03}   & \multicolumn{1}{c}{81.61}   & \multicolumn{1}{c}{89.24}    & \multicolumn{1}{c}{54.24} & 228.88 \\ \hline
6 & \multicolumn{1}{c}{$\checkmark$}    &   $\checkmark$   & \multicolumn{1}{c}{$\checkmark$}    &          $\checkmark$                & \multicolumn{1}{c}{$\checkmark$}    &        $\checkmark$                  & \multicolumn{1}{c|}{$\checkmark$} & \multicolumn{1}{c}{58.47}   & \multicolumn{1}{c}{82.01}   & \multicolumn{1}{c}{89.52}    & \multicolumn{1}{c}{54.59} & 230.00 \\ \hline
\end{tabular}
}
\end{table*}

\begin{table}[htbp]
\caption{Quantitative comparison of different modality fusion strategies. Cat denotes feature concatenation, Add represents element-wise addition, and ATF refers to our proposed Adaptive Token Fusion module.~\label{tab:table5}}
\renewcommand{\arraystretch}{1.3}
\setlength{\tabcolsep}{0.2cm}{
\centering
\begin{tabular}{c|ccccc}
\hline
Fusion & Rank-1 & Rank-5 & Rank-10 & mAP & RSum \\
\hline
Cat & 50.48 & 77.91 & 85.58 & 48.56 & 213.97 \\
Add & 58.03 & 81.61 & 89.23 & 54.24 & 228.87 \\
ATF & 58.47 & 82.01 & 89.52 & 54.59  & 230.00 \\
\hline
\end{tabular}
}
\end{table}

\noindent\textbf{Effectiveness of Modality-Collaborative Visual Text Alignment.}
As shown in Table~\ref{tab:table4}, we conduct an in-depth ablation study on the modality-collaborative visual-text alignment module to investigate the impact of different loss functions under the setting where only RGBT and text are aligned. 
Specifically, Experiment No.0 represents the baseline method we adopt as the reference, while Experiment No.1 introduces the Arbitrary-word Reconstruction (AR) loss on top of the baseline.
The results show that introducing the AR loss improves Rank-1 and mAP by 0.4\% and 0.43\%, respectively, indicating that this loss helps enhance the alignment between RGBT multimodal representations and textual features. 
However, it is worth noting that relying solely on the modality-cooperative alignment strategy still leads to limited retrieval performance, mainly because it fails to fully exploit the complementary advantages of different visual modalities in the semantic alignment process.

\noindent\textbf{Effectiveness of Modality-Specific Visual Text Alignment.}
Table~\ref{tab:table4} (Experiments No.2–5) evaluates the effectiveness of different loss functions within the modality-specific visual-text alignment framework. 
The results demonstrate that, compared to modality-cooperative alignment, modality-specific alignment further enhances the association between RGBT multimodal representations and text. Specifically, comparing No.1 and No.5, the inclusion of modality-specific visual-text alignment leads to a 2.73\% improvement in Rank-1 and a 1.62\% improvement in mAP, clearly validating its contribution to more accurate RGBT-text alignment.
Comparing No.1 and No.2, the addition of the BIA module results in a 1.33\% gain in Rank-1, while comparing No.3 and No.2, the inclusion of the CRS module brings an additional 0.57\% improvement in Rank-1. These findings highlight not only the benefit of aligning RGB with text but also confirm the effectiveness of each component.
Further, comparing No.4 and No.3, the proposed UIB module boosts Rank-1 by 0.76\%, and comparing No.5 and No.4, the introduction of CUS yields a 1.58\% gain in RSum. These improvements confirm that aligning the T modality with text contributes positively to the overall alignment performance and verify the utility of each module.

\noindent\textbf{Effectiveness of Adaptive Token Fusion (ATF).}
Comparing No.6 with No.5 in Table~\ref{tab:table4}, we observe a 0.44\% increase in Rank-1 and a 0.35\% increase in mAP, with the overall RSum reaching 230.00\%, which demonstrates the effectiveness of the proposed ATF module.
This indicates that ATF successfully generates more precise RGBT fusion embeddings by predicting convex combination weights for the two visual modalities and combining them with enhanced mixed visual embeddings.
Moreover, in Table~\ref{tab:table5}, we compare ATF with the additive fusion and concatenation fusion methods. 
The results show that our proposed ATF significantly outperforms both of these fusion methods across all metrics.

\begin{table}[htbp]
\caption{Experiments on different combinations of text masking methods on the RGBT-PEDES benchmark. RM denotes random mask, CRM denotes color-related mask and CUM represents color-unrelated mask. ~\label{tab:table6}}
\renewcommand{\arraystretch}{1.3}
\setlength{\tabcolsep}{1.1pt}
\setlength{\tabcolsep}{0.1cm}{
\centering
\begin{tabular}{l|cccc}
\hline
Methods & Rank-1 & Rank-5 & Rank-10 & mAP \\
\hline
RM-RM-RM & 57.75  & 81.44 & 89.19  & 53.90 \\
RM-RM-CUM & 57.07 & 81.04 & 88.87 & 54.08 \\
CRM-RM-RM & 57.47 & 81.41 & 89.07 & 54.18 \\
CRM-RM-CUM (Ours) & 58.47 & 82.01 & 89.52 & 54.59  \\
\hline
\end{tabular}
}
\end{table}
\noindent\textbf{Effectiveness of Modality Specific Text Mask.}
To further validate the effectiveness of modality-specific text masks, we conducted experiments with various combinations of masking methods.
As shown in Table~\ref{tab:table6}, we tested different mask combinations, including complete random mask (RM-RM-RM), random mask - random mask - color unrelated mask (RM-RM-CUM), color-related mask - random mask - random mask (CRM-RM-RM), and our proposed modality-specific mask (CRM-RM-CUM).
Compared to other mask combinations, our approach outperformed all other methods across all evaluation metrics, demonstrating significant advantages.
This fully confirms the applicability and effectiveness of our proposed modality-specific text masks for the current task.

\begin{figure}[!t]
\centering
\includegraphics[width=3in,height=1.6in]{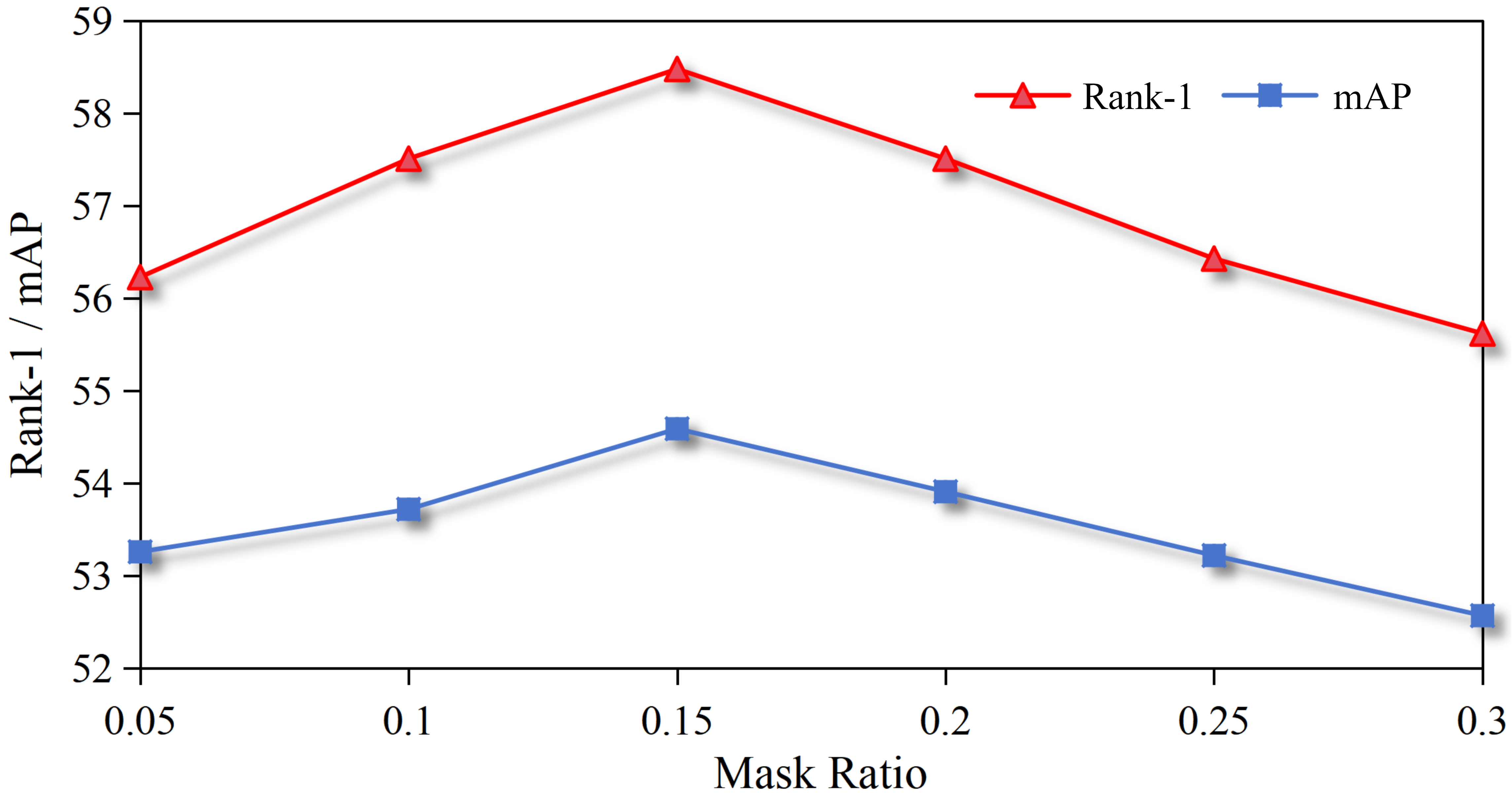}
\caption{Quantitative comparison under different mask ratios.
}
\label{fig_8}
\end{figure}
\subsection{Parameter sensitivity analysis}
To study the impact of the text masking ratio on model performance, we conduct a sensitivity analysis on the RGBT-PEDES dataset. As shown in Figure~\ref{fig_8}, we observe that: (1) When the masking ratio increases from 0.05 to 0.15, both Rank-1 and mAP improve steadily. (2) However, further increasing the masking ratio beyond 0.15 results in a consistent decline in performance. We achieve the best performance (58.47\% Rank-1 and 54.59\% mAP) when the masking ratio is set to 0.15, which we adopt in all our experiments.
We attribute this pattern to the balance between retaining sufficient textual information and encouraging fine-grained semantic alignment. A small masking ratio (e.g., 0.05) leaves most of the text unchanged, providing little challenge for the model and limiting its ability to learn robust associations under partial descriptions. Conversely, a large masking ratio (e.g., 0.25 or higher) removes too much semantic content, making it difficult for the model to establish meaningful alignments, especially in scenarios that rely on detailed attribute descriptions. Therefore, a moderate masking ratio provides a better trade-off between generalization and discrimination.

\subsection{Comparison of Model Efficiency}
\begin{table}[htbp]
\renewcommand{\arraystretch}{1.3}
\setlength{\tabcolsep}{0.1cm}{
\centering
\begin{tabular}{c|cccc}
\hline
 & FLOPs & Train Time & Inference Speed & Rank-1 \\
\hline
APTM & 59.3(G) & 0.9(h) & 35.9(ms)  & 53.13 \\
IRRA & 24.0(G) & 1.1(h) & 24.7(ms) & 55.62 \\
CFAM & 39.1(G) & 0.6(h) & 28.7(ms) & 55.46 \\
Ours & 26.0(G) & 0.9(h) & 26.7(ms) & 58.47 \\
\hline
\end{tabular}
}
\end{table}
\begin{figure}[!t]
\centering
\includegraphics[width=3.1in,height=3.9in]{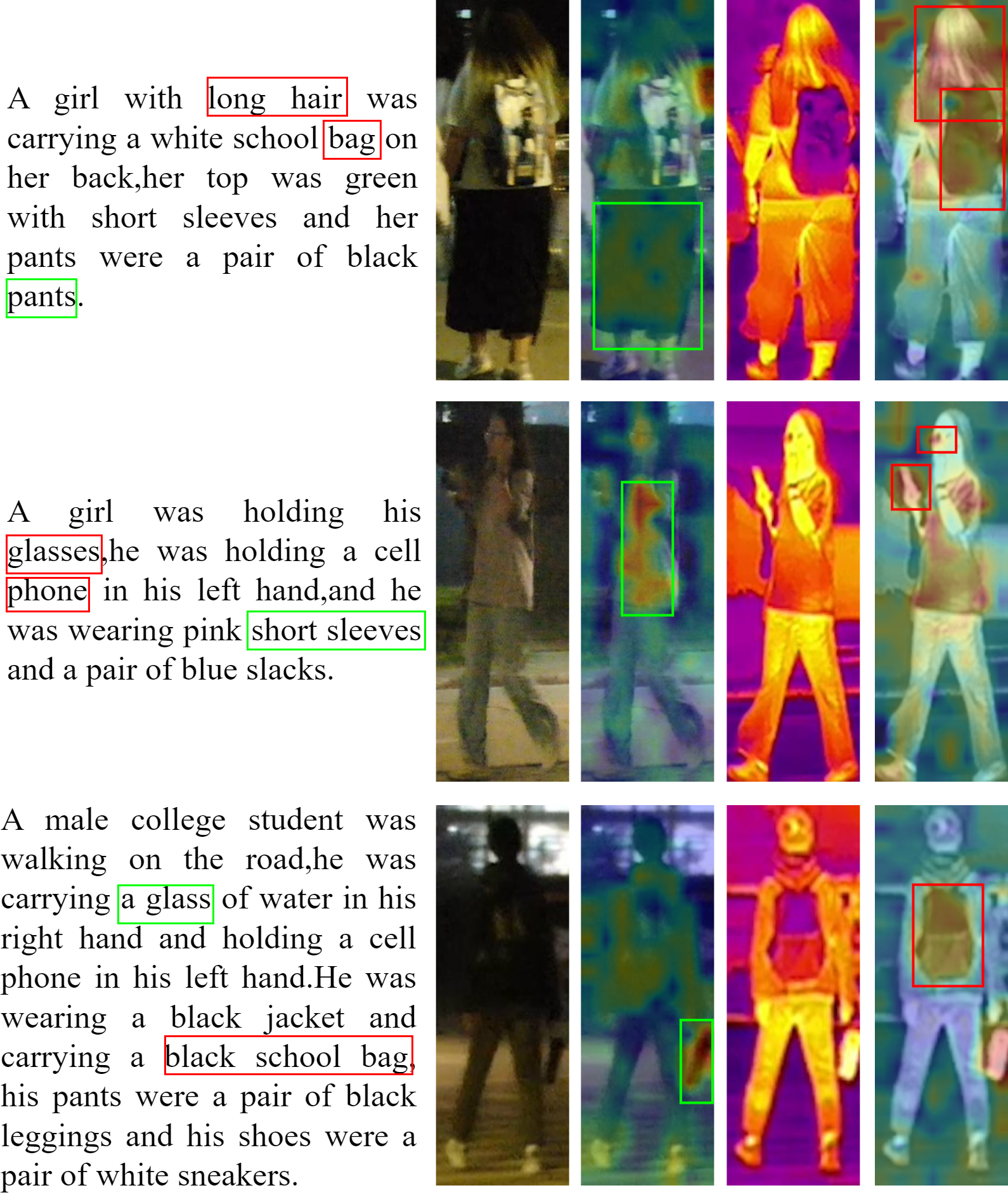}
\caption{Visualization of attention between text and RGB/T images. In each row, the first column shows the input text; the second and third columns present the RGB image and its overlaid attention map; the fourth and fifth columns display the T image and its corresponding attention map. Words highlighted with green bounding boxes in the text correspond to attended regions outlined in green in the RGB attention maps, while red-highlighted words indicate corresponding attention regions in the thermal maps.}
\label{fig_13}
\end{figure}
\begin{figure}[htbp]
\centering\includegraphics[width=3.3in,height=3in]{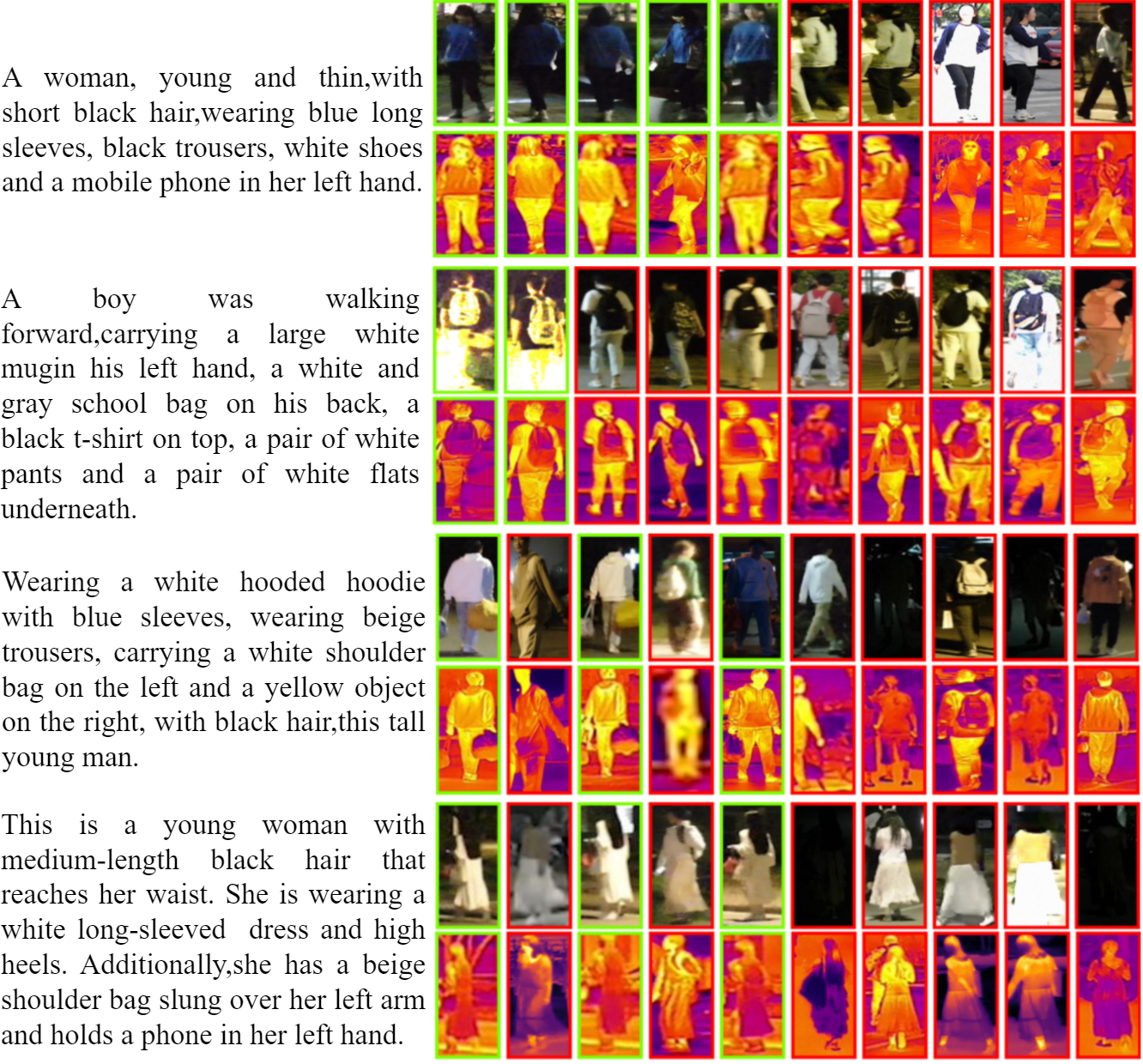}
\caption{Qualitative retrieval results generated by our method on the RGBT-PEDES dataset. Green bounding boxes indicate correct matches, while red bounding boxes indicate incorrect ones.}
\label{fig_14}
\end{figure}
We conduct a comprehensive comparison of DCAlign with three state-of-the-art models: IRRA, APTM, and CFAM.
Compared to APTM, our method reduces FLOPS by 56.2\%, significantly improving computational efficiency, while also increasing speed by 25.6\%, achieving faster processing times without sacrificing performance. 
In comparison to IRRA, our method improves Rank-1 accuracy (R-1) by 2.8\%, demonstrating enhanced recognition capabilities. 
When compared with the CFAM model, our method reduces FLOPS by approximately 50.4\%, greatly lowering computational resource consumption. At the same time, our method achieves a 5.4\% improvement in Rank-1 accuracy. 
Although training time increases by 50\%, our inference speed improves by approximately 7\%, resulting in faster inference and lower latency.
Overall, our method significantly improves accuracy while enhancing computational efficiency and inference speed, showcasing an excellent balance between resource utilization and model performance.
Compared to existing methods, our approach not only optimizes efficiency but also improves the overall performance of the model.

\subsection{Qualitative Attention Study} 
To gain a deeper understanding of how the model performs alignment between text and images, we conduct a qualitative analysis of its cross-modal attention. 
Specifically, we visualize the attention weights from textual queries to both RGB and T image features, as shown in the Figure~\ref{fig_13}.
The figure presents three representative textual descriptions, with attention maps overlaid on the corresponding RGB and T images. 
Keywords marked with green boxes in the text correspond to attention regions highlighted in green on the RGB images, while those marked with red boxes correspond to red-highlighted regions on the thermal images. 
For better readability, the thermal images are rendered in grayscale and superimposed with heatmaps to enhance the visibility of the attention regions.
In the first example, due to poor lighting conditions, the RGB image fails to provide sufficient cues for identifying “long hair,” while the thermal image reveals clearer human contours, enabling the model to more accurately locate the referenced region. 
Meanwhile, for the description “black pants,” the RGB image retains richer texture details, resulting in stronger attention in that region under the RGB modality.
In the second example, “glasses” and “phone” exhibit stronger heat signatures in the thermal image, leading the model to allocate more attention in the T modality. 
In contrast, “sleeves” are better represented in the RGB image in terms of texture, thus receiving more focused attention in the RGB modality.
In the third example, the shape of the “school bag” is more clearly distinguishable in the thermal image, allowing the model to establish a more precise semantic correspondence in that modality.
These visualizations confirm that the proposed model adaptively attends to semantically relevant regions across modalities. 
It effectively captures the fine-grained appearance details from RGB images and the stable contour cues from thermal images, thereby achieving more accurate and robust cross-modal semantic alignment.

\subsection{Qualitative Analysis of Retrieval Results}
To further validate the effectiveness of our method, we present additional qualitative retrieval results on the RGBT-PEDES dataset, showcasing the real-world performance of model under diverse query texts and challenging environments.
As shown in Figure~\ref{fig_14}, our method consistently retrieves semantically relevant person images, even under complex lighting conditions such as low-light and overexposure.
The top-10 retrieval results demonstrate that our model accurately understands fine-grained semantic cues in the text and effectively aligns them with visual attributes across the visible, thermal, and fused modalities. 
Specifically, the model successfully attends to key textual descriptions, such as clothing colors, carried objects, and accessories, enabling it to distinguish visually similar yet semantically different individuals.
These findings highlight the strength of our proposed multi-level global-local alignment mechanism and adaptive fusion strategy, which significantly enhance the capability of model in cross-modal semantic modeling and discrimination.
The qualitative examples not only reinforce the quantitative superiority of our method but also demonstrate its interpretability and practical applicability in real-world scenarios, making it well-suited for robust text-based person retrieval under complex environmental conditions.

\section{Conclusion}
We introduce the Text-RGBT person retrieval task, a novel and challenging problem that integrates textual descriptions with both visible (RGB) and thermal (T) modalities for robust person retrieval under complex real-world conditions.
To tackle this task, we propose DCAlign, a cross-modal alignment framework that explicitly decouples semantic attributes and jointly models both modality-collaborative and modality-specific visual-text alignment. By constructing a multi-level image-text association mechanism, DCAlign aligns cross-modal semantics at both global and local levels, effectively leveraging the complementary strengths of RGB and T modalities for accurate and resilient retrieval.
We also present RGBT-PEDES, a high-quality benchmark dataset specifically designed for this task. It reflects real-world complexities such as varying illumination, cluttered backgrounds, and diverse scenes, providing a solid foundation for training and evaluation.
Extensive experiments on RGBT-PEDES demonstrate that DCAlign significantly outperforms existing methods in terms of Rank-1, mAP, and other metrics. It effectively mitigates cross-modal alignment gaps while capturing fine-grained semantic details, validating the effectiveness of DCAlign and its potential to advance RGBT based cross-modal retrieval research.

In the future, Text-RGBT person retrieval can be further advanced along several key directions.
First, constructing larger-scale and more diverse annotated datasets is essential to improve the generalization and real-world applicability of retrieval models. 
Techniques such as synthetic data generation, automatic annotation, and cross-domain augmentation may help alleviate current data limitations.
Second, exploring cross-modal pretraining strategies based on vision-language foundation models (e.g., CLIP, ImageBind) adapted to the RGBT domain could greatly enhance semantic alignment and robustness.
Third, fully exploiting the complementary advantages of thermal modality, especially under low-light or adverse conditions, remains a promising direction. This includes designing more explicit and interpretable mechanisms to emphasize thermal cues in the feature learning process.
Finally, developing lightweight and efficient architectures is crucial for real-world deployment, particularly in resource-constrained environments. Efficient model design can enable faster inference and broader applicability on edge devices.


{\flushleft\bf Data Availability Statement.}

The data used to support the findings of this study are included in the paper.

{\flushleft\bf Declaration of Competing Interest.}

No potential conflict of interest was reported by the authors.

{\flushleft\bf Author Statement.}

Yifei Deng: Conceptualization of this study, Methodology, Writing – original draft, and Software. Chenglong Li: Project administration, Funding acquisition, Investigation, and Writing     – review and editing. Zhenyu Chen: Data curation, Validation. Zihen Xu: Visualization and Supervision. Jin Tang: Resources, Formal analysis, Interpretation of data, and Writing – review and editing.

{\flushleft\bf Acknowledgements.}
This research is partly supported by the National Natural Science Foundation of China (No. 62376004) and the Natural Science Foundation of Anhui Province (No. 2208085J18).

\bibliographystyle{cas-model2-names}
\bibliography{cas-refs}

\begin{thebibliography}{51}
\expandafter\ifx\csname natexlab\endcsname\relax\def\natexlab#1{#1}\fi
\providecommand{\url}[1]{\texttt{#1}}
\providecommand{\href}[2]{#2}
\providecommand{\path}[1]{#1}
\providecommand{\DOIprefix}{doi:}
\providecommand{\ArXivprefix}{arXiv:}
\providecommand{\URLprefix}{URL: }
\providecommand{\Pubmedprefix}{pmid:}
\providecommand{\doi}[1]{\href{http://dx.doi.org/#1}{\path{#1}}}
\providecommand{\Pubmed}[1]{\href{pmid:#1}{\path{#1}}}
\providecommand{\bibinfo}[2]{#2}
\ifx\xfnm\relax \def\xfnm[#1]{\unskip,\space#1}\fi
\bibitem[{Cao et~al.(2024)Cao, Bai, Zeng, Ye and Zhang}]{cao2024empirical}
\bibinfo{author}{Cao, M.}, \bibinfo{author}{Bai, Y.}, \bibinfo{author}{Zeng, Z.}, \bibinfo{author}{Ye, M.}, \bibinfo{author}{Zhang, M.}, \bibinfo{year}{2024}.
\newblock \bibinfo{title}{An empirical study of clip for text-based person search}, in: \bibinfo{booktitle}{Proceedings of the AAAI Conference on Artificial Intelligence}, pp. \bibinfo{pages}{465--473}.
\bibitem[{Chen et~al.(2018)Chen, Xu and Luo}]{chen2018improving}
\bibinfo{author}{Chen, T.}, \bibinfo{author}{Xu, C.}, \bibinfo{author}{Luo, J.}, \bibinfo{year}{2018}.
\newblock \bibinfo{title}{Improving text-based person search by spatial matching and adaptive threshold}, in: \bibinfo{booktitle}{2018 IEEE Winter Conference on Applications of Computer Vision (WACV)}, \bibinfo{organization}{IEEE}. pp. \bibinfo{pages}{1879--1887}.
\bibitem[{Chen et~al.(2021)Chen, Huang, Chang, Tan, Xue and Ma}]{chen2021cross}
\bibinfo{author}{Chen, Y.}, \bibinfo{author}{Huang, R.}, \bibinfo{author}{Chang, H.}, \bibinfo{author}{Tan, C.}, \bibinfo{author}{Xue, T.}, \bibinfo{author}{Ma, B.}, \bibinfo{year}{2021}.
\newblock \bibinfo{title}{Cross-modal knowledge adaptation for language-based person search}.
\newblock \bibinfo{journal}{IEEE Transactions on Image Processing} \bibinfo{volume}{30}, \bibinfo{pages}{4057--4069}.
\bibitem[{Chen et~al.(2022)Chen, Zhang, Lu, Wang and Zheng}]{chen2022tipcb}
\bibinfo{author}{Chen, Y.}, \bibinfo{author}{Zhang, G.}, \bibinfo{author}{Lu, Y.}, \bibinfo{author}{Wang, Z.}, \bibinfo{author}{Zheng, Y.}, \bibinfo{year}{2022}.
\newblock \bibinfo{title}{Tipcb: A simple but effective part-based convolutional baseline for text-based person search}.
\newblock \bibinfo{journal}{Neurocomputing} \bibinfo{volume}{494}, \bibinfo{pages}{171--181}.
\bibitem[{Conaire et~al.(2006)Conaire, O'Connor, Cooke and Smeaton}]{conaire2006comparison}
\bibinfo{author}{Conaire, C.O.}, \bibinfo{author}{O'Connor, N.E.}, \bibinfo{author}{Cooke, E.}, \bibinfo{author}{Smeaton, A.F.}, \bibinfo{year}{2006}.
\newblock \bibinfo{title}{Comparison of fusion methods for thermo-visual surveillance tracking}, in: \bibinfo{booktitle}{2006 9th International Conference on Information Fusion}, \bibinfo{organization}{IEEE}. pp. \bibinfo{pages}{1--7}.
\bibitem[{Dai et~al.(2025)Dai, Xu, Li, Zhang and Xia}]{dai2025humanvlm}
\bibinfo{author}{Dai, D.}, \bibinfo{author}{Xu, L.}, \bibinfo{author}{Li, Y.}, \bibinfo{author}{Zhang, Y.}, \bibinfo{author}{Xia, S.}, \bibinfo{year}{2025}.
\newblock \bibinfo{title}{Humanvlm: Foundation for human-scene vision-language model}.
\newblock \bibinfo{journal}{Information Fusion} , \bibinfo{pages}{103271}.
\bibitem[{Deng et~al.(2025)Deng, Chen, Li and Tang}]{deng2025uncertainty}
\bibinfo{author}{Deng, Y.}, \bibinfo{author}{Chen, Z.}, \bibinfo{author}{Li, C.}, \bibinfo{author}{Tang, J.}, \bibinfo{year}{2025}.
\newblock \bibinfo{title}{Uncertainty-aware coarse-to-fine alignment for text-image person retrieval}.
\newblock \bibinfo{journal}{Visual Intelligence} \bibinfo{volume}{3}, \bibinfo{pages}{6}.
\bibitem[{Deng et~al.(2022)Deng, Li, Lu, Li and Luo}]{deng2022factory}
\bibinfo{author}{Deng, Y.}, \bibinfo{author}{Li, C.}, \bibinfo{author}{Lu, A.}, \bibinfo{author}{Li, W.}, \bibinfo{author}{Luo, B.}, \bibinfo{year}{2022}.
\newblock \bibinfo{title}{Factory extraction from satellite images: Benchmark and baseline}.
\newblock \bibinfo{journal}{Remote Sensing} \bibinfo{volume}{14}, \bibinfo{pages}{5657}.
\bibitem[{Deng et~al.(2024)Deng, Wang, Li, Wang, Zhang and Tang}]{deng2024collaborative}
\bibinfo{author}{Deng, Y.}, \bibinfo{author}{Wang, G.}, \bibinfo{author}{Li, C.}, \bibinfo{author}{Wang, W.}, \bibinfo{author}{Zhang, C.}, \bibinfo{author}{Tang, J.}, \bibinfo{year}{2024}.
\newblock \bibinfo{title}{Collaborative license plate recognition via association enhancement network with auxiliary learning and a unified benchmark}.
\newblock \bibinfo{journal}{IEEE Transactions on Multimedia} .
\bibitem[{Feng and Su(2024)}]{feng2024rgbt}
\bibinfo{author}{Feng, M.}, \bibinfo{author}{Su, J.}, \bibinfo{year}{2024}.
\newblock \bibinfo{title}{Rgbt tracking: A comprehensive review}.
\newblock \bibinfo{journal}{Information Fusion} , \bibinfo{pages}{102492}.
\bibitem[{Gao et~al.(2021)Gao, Cai, Jiang, Zheng, Zhang, Gong, Peng, Guo and Sun}]{gao2021contextual}
\bibinfo{author}{Gao, C.}, \bibinfo{author}{Cai, G.}, \bibinfo{author}{Jiang, X.}, \bibinfo{author}{Zheng, F.}, \bibinfo{author}{Zhang, J.}, \bibinfo{author}{Gong, Y.}, \bibinfo{author}{Peng, P.}, \bibinfo{author}{Guo, X.}, \bibinfo{author}{Sun, X.}, \bibinfo{year}{2021}.
\newblock \bibinfo{title}{Contextual non-local alignment over full-scale representation for text-based person search}.
\newblock \bibinfo{journal}{arXiv preprint arXiv:2101.03036} .
\bibitem[{Gong et~al.(2024)Gong, Wang and Zhang}]{gong2024cross}
\bibinfo{author}{Gong, T.}, \bibinfo{author}{Wang, J.}, \bibinfo{author}{Zhang, L.}, \bibinfo{year}{2024}.
\newblock \bibinfo{title}{Cross-modal semantic aligning and neighbor-aware completing for robust text--image person retrieval}.
\newblock \bibinfo{journal}{Information Fusion} \bibinfo{volume}{112}, \bibinfo{pages}{102544}.
\bibitem[{Ha et~al.(2017)Ha, Watanabe, Karasawa, Ushiku and Harada}]{ha2017mfnet}
\bibinfo{author}{Ha, Q.}, \bibinfo{author}{Watanabe, K.}, \bibinfo{author}{Karasawa, T.}, \bibinfo{author}{Ushiku, Y.}, \bibinfo{author}{Harada, T.}, \bibinfo{year}{2017}.
\newblock \bibinfo{title}{Mfnet: Towards real-time semantic segmentation for autonomous vehicles with multi-spectral scenes}, in: \bibinfo{booktitle}{2017 IEEE/RSJ International Conference on Intelligent Robots and Systems (IROS)}, \bibinfo{organization}{IEEE}. pp. \bibinfo{pages}{5108--5115}.
\bibitem[{Han et~al.(2021)Han, He, Zhang and Xiang}]{han2021text}
\bibinfo{author}{Han, X.}, \bibinfo{author}{He, S.}, \bibinfo{author}{Zhang, L.}, \bibinfo{author}{Xiang, T.}, \bibinfo{year}{2021}.
\newblock \bibinfo{title}{Text-based person search with limited data}.
\newblock \bibinfo{journal}{arXiv preprint arXiv:2110.10807} .
\bibitem[{Hwang et~al.(2015)Hwang, Park, Kim, Choi and So~Kweon}]{hwang2015multispectral}
\bibinfo{author}{Hwang, S.}, \bibinfo{author}{Park, J.}, \bibinfo{author}{Kim, N.}, \bibinfo{author}{Choi, Y.}, \bibinfo{author}{So~Kweon, I.}, \bibinfo{year}{2015}.
\newblock \bibinfo{title}{Multispectral pedestrian detection: Benchmark dataset and baseline}, in: \bibinfo{booktitle}{Proceedings of the IEEE conference on computer vision and pattern recognition}, pp. \bibinfo{pages}{1037--1045}.
\bibitem[{Ji et~al.(2023)Ji, Li, Bian, Zhou, Zhao, Yuille and Cheng}]{ji2023multispectral}
\bibinfo{author}{Ji, W.}, \bibinfo{author}{Li, J.}, \bibinfo{author}{Bian, C.}, \bibinfo{author}{Zhou, Z.}, \bibinfo{author}{Zhao, J.}, \bibinfo{author}{Yuille, A.L.}, \bibinfo{author}{Cheng, L.}, \bibinfo{year}{2023}.
\newblock \bibinfo{title}{Multispectral video semantic segmentation: A benchmark dataset and baseline}, in: \bibinfo{booktitle}{Proceedings of the IEEE/CVF Conference on Computer Vision and Pattern Recognition}, pp. \bibinfo{pages}{1094--1104}.
\bibitem[{Jiang and Ye(2023)}]{jiang2023cross}
\bibinfo{author}{Jiang, D.}, \bibinfo{author}{Ye, M.}, \bibinfo{year}{2023}.
\newblock \bibinfo{title}{Cross-modal implicit relation reasoning and aligning for text-to-image person retrieval}, in: \bibinfo{booktitle}{Proceedings of the IEEE/CVF Conference on Computer Vision and Pattern Recognition}, pp. \bibinfo{pages}{2787--2797}.
\bibitem[{Jiang et~al.(2025)Jiang, Yang, Jones and Zhang}]{jiang2025attributes}
\bibinfo{author}{Jiang, F.}, \bibinfo{author}{Yang, S.}, \bibinfo{author}{Jones, M.W.}, \bibinfo{author}{Zhang, L.}, \bibinfo{year}{2025}.
\newblock \bibinfo{title}{From attributes to natural language: A survey and foresight on text-based person re-identification}.
\newblock \bibinfo{journal}{Information Fusion} \bibinfo{volume}{118}, \bibinfo{pages}{102879}.
\bibitem[{Jing et~al.(2020)Jing, Si, Wang, Wang, Wang and Tan}]{jing2020pose}
\bibinfo{author}{Jing, Y.}, \bibinfo{author}{Si, C.}, \bibinfo{author}{Wang, J.}, \bibinfo{author}{Wang, W.}, \bibinfo{author}{Wang, L.}, \bibinfo{author}{Tan, T.}, \bibinfo{year}{2020}.
\newblock \bibinfo{title}{Pose-guided multi-granularity attention network for text-based person search}, in: \bibinfo{booktitle}{Proceedings of the AAAI Conference on Artificial Intelligence}, pp. \bibinfo{pages}{11189--11196}.
\bibitem[{Li et~al.(2019)Li, Liang, Lu, Zhao and Tang}]{li2019rgb}
\bibinfo{author}{Li, C.}, \bibinfo{author}{Liang, X.}, \bibinfo{author}{Lu, Y.}, \bibinfo{author}{Zhao, N.}, \bibinfo{author}{Tang, J.}, \bibinfo{year}{2019}.
\newblock \bibinfo{title}{Rgb-t object tracking: Benchmark and baseline}.
\newblock \bibinfo{journal}{Pattern Recognition} \bibinfo{volume}{96}, \bibinfo{pages}{106977}.
\bibitem[{Li et~al.(2021)Li, Xue, Jia, Qu, Luo, Tang and Sun}]{li2021lasher}
\bibinfo{author}{Li, C.}, \bibinfo{author}{Xue, W.}, \bibinfo{author}{Jia, Y.}, \bibinfo{author}{Qu, Z.}, \bibinfo{author}{Luo, B.}, \bibinfo{author}{Tang, J.}, \bibinfo{author}{Sun, D.}, \bibinfo{year}{2021}.
\newblock \bibinfo{title}{Lasher: A large-scale high-diversity benchmark for rgbt tracking}.
\newblock \bibinfo{journal}{IEEE Transactions on Image Processing} \bibinfo{volume}{31}, \bibinfo{pages}{392--404}.
\bibitem[{Li et~al.(2025)Li, Chen, Wang, Zhong and Xiao}]{li2025adapting}
\bibinfo{author}{Li, J.}, \bibinfo{author}{Chen, T.}, \bibinfo{author}{Wang, X.}, \bibinfo{author}{Zhong, Y.}, \bibinfo{author}{Xiao, X.}, \bibinfo{year}{2025}.
\newblock \bibinfo{title}{Adapting the segment anything model for multi-modal retinal anomaly detection and localization}.
\newblock \bibinfo{journal}{Information Fusion} \bibinfo{volume}{113}, \bibinfo{pages}{102631}.
\bibitem[{Li et~al.(2017)Li, Xiao, Li, Zhou, Yue and Wang}]{li2017person}
\bibinfo{author}{Li, S.}, \bibinfo{author}{Xiao, T.}, \bibinfo{author}{Li, H.}, \bibinfo{author}{Zhou, B.}, \bibinfo{author}{Yue, D.}, \bibinfo{author}{Wang, X.}, \bibinfo{year}{2017}.
\newblock \bibinfo{title}{Person search with natural language description}, in: \bibinfo{booktitle}{Proceedings of the IEEE conference on computer vision and pattern recognition}, pp. \bibinfo{pages}{1970--1979}.
\bibitem[{Lin et~al.(2024)Lin, Yin, Ping, Molchanov, Shoeybi and Han}]{lin2024vila}
\bibinfo{author}{Lin, J.}, \bibinfo{author}{Yin, H.}, \bibinfo{author}{Ping, W.}, \bibinfo{author}{Molchanov, P.}, \bibinfo{author}{Shoeybi, M.}, \bibinfo{author}{Han, S.}, \bibinfo{year}{2024}.
\newblock \bibinfo{title}{Vila: On pre-training for visual language models}, in: \bibinfo{booktitle}{Proceedings of the IEEE/CVF Conference on Computer Vision and Pattern Recognition}, pp. \bibinfo{pages}{26689--26699}.
\bibitem[{Liu et~al.(2021)Liu, Chen, Wu, Li, Li and Lin}]{liu2021cross}
\bibinfo{author}{Liu, L.}, \bibinfo{author}{Chen, J.}, \bibinfo{author}{Wu, H.}, \bibinfo{author}{Li, G.}, \bibinfo{author}{Li, C.}, \bibinfo{author}{Lin, L.}, \bibinfo{year}{2021}.
\newblock \bibinfo{title}{Cross-modal collaborative representation learning and a large-scale rgbt benchmark for crowd counting}, in: \bibinfo{booktitle}{Proceedings of the IEEE/CVF conference on computer vision and pattern recognition}, pp. \bibinfo{pages}{4823--4833}.
\bibitem[{Mohammed and Kora(2023)}]{mohammed2023comprehensive}
\bibinfo{author}{Mohammed, A.}, \bibinfo{author}{Kora, R.}, \bibinfo{year}{2023}.
\newblock \bibinfo{title}{A comprehensive review on ensemble deep learning: Opportunities and challenges}.
\newblock \bibinfo{journal}{Journal of King Saud University-Computer and Information Sciences} \bibinfo{volume}{35}, \bibinfo{pages}{757--774}.
\bibitem[{Niu et~al.(2020)Niu, Huang, Ouyang and Wang}]{niu2020improving}
\bibinfo{author}{Niu, K.}, \bibinfo{author}{Huang, Y.}, \bibinfo{author}{Ouyang, W.}, \bibinfo{author}{Wang, L.}, \bibinfo{year}{2020}.
\newblock \bibinfo{title}{Improving description-based person re-identification by multi-granularity image-text alignments}.
\newblock \bibinfo{journal}{IEEE Transactions on Image Processing} \bibinfo{volume}{29}, \bibinfo{pages}{5542--5556}.
\bibitem[{Quan et~al.(2025)Quan, Hou, Yin and Zhang}]{quan2025multi}
\bibinfo{author}{Quan, X.}, \bibinfo{author}{Hou, G.}, \bibinfo{author}{Yin, W.}, \bibinfo{author}{Zhang, H.}, \bibinfo{year}{2025}.
\newblock \bibinfo{title}{A multi-modal and multi-stage fusion enhancement network for segmentation based on oct and octa images}.
\newblock \bibinfo{journal}{Information Fusion} \bibinfo{volume}{113}, \bibinfo{pages}{102594}.
\bibitem[{Radford et~al.(2021)Radford, Kim, Hallacy, Ramesh, Goh, Agarwal, Sastry, Askell, Mishkin, Clark et~al.}]{radford2021learning}
\bibinfo{author}{Radford, A.}, \bibinfo{author}{Kim, J.W.}, \bibinfo{author}{Hallacy, C.}, \bibinfo{author}{Ramesh, A.}, \bibinfo{author}{Goh, G.}, \bibinfo{author}{Agarwal, S.}, \bibinfo{author}{Sastry, G.}, \bibinfo{author}{Askell, A.}, \bibinfo{author}{Mishkin, P.}, \bibinfo{author}{Clark, J.}, et~al., \bibinfo{year}{2021}.
\newblock \bibinfo{title}{Learning transferable visual models from natural language supervision}, in: \bibinfo{booktitle}{International conference on machine learning}, \bibinfo{organization}{PmLR}. pp. \bibinfo{pages}{8748--8763}.
\bibitem[{Ring and Ammer(2012)}]{ring2012infrared}
\bibinfo{author}{Ring, E.}, \bibinfo{author}{Ammer, K.}, \bibinfo{year}{2012}.
\newblock \bibinfo{title}{Infrared thermal imaging in medicine}.
\newblock \bibinfo{journal}{Physiological measurement} \bibinfo{volume}{33}, \bibinfo{pages}{R33}.
\bibitem[{Sarafianos et~al.(2019)Sarafianos, Xu and Kakadiaris}]{sarafianos2019adversarial}
\bibinfo{author}{Sarafianos, N.}, \bibinfo{author}{Xu, X.}, \bibinfo{author}{Kakadiaris, I.A.}, \bibinfo{year}{2019}.
\newblock \bibinfo{title}{Adversarial representation learning for text-to-image matching}, in: \bibinfo{booktitle}{Proceedings of the IEEE/CVF international conference on computer vision}, pp. \bibinfo{pages}{5814--5824}.
\bibitem[{Shao et~al.(2022)Shao, Zhang, Fang, Lin, Wang and Ding}]{shao2022learning}
\bibinfo{author}{Shao, Z.}, \bibinfo{author}{Zhang, X.}, \bibinfo{author}{Fang, M.}, \bibinfo{author}{Lin, Z.}, \bibinfo{author}{Wang, J.}, \bibinfo{author}{Ding, C.}, \bibinfo{year}{2022}.
\newblock \bibinfo{title}{Learning granularity-unified representations for text-to-image person re-identification}, in: \bibinfo{booktitle}{Proceedings of the 30th acm international conference on multimedia}, pp. \bibinfo{pages}{5566--5574}.
\bibitem[{Sun et~al.(2022)Sun, Cao, Zhu and Hu}]{sun2022drone}
\bibinfo{author}{Sun, Y.}, \bibinfo{author}{Cao, B.}, \bibinfo{author}{Zhu, P.}, \bibinfo{author}{Hu, Q.}, \bibinfo{year}{2022}.
\newblock \bibinfo{title}{Drone-based rgb-infrared cross-modality vehicle detection via uncertainty-aware learning}.
\newblock \bibinfo{journal}{IEEE Transactions on Circuits and Systems for Video Technology} \bibinfo{volume}{32}, \bibinfo{pages}{6700--6713}.
\bibitem[{Tang et~al.(2019)Tang, Fan, Wang, Tu and Li}]{tang2019rgbt}
\bibinfo{author}{Tang, J.}, \bibinfo{author}{Fan, D.}, \bibinfo{author}{Wang, X.}, \bibinfo{author}{Tu, Z.}, \bibinfo{author}{Li, C.}, \bibinfo{year}{2019}.
\newblock \bibinfo{title}{Rgbt salient object detection: Benchmark and a novel cooperative ranking approach}.
\newblock \bibinfo{journal}{IEEE Transactions on Circuits and Systems for Video Technology} \bibinfo{volume}{30}, \bibinfo{pages}{4421--4433}.
\bibitem[{Tu et~al.(2022)Tu, Ma, Li, Li, Xu and Liu}]{tu2022rgbt}
\bibinfo{author}{Tu, Z.}, \bibinfo{author}{Ma, Y.}, \bibinfo{author}{Li, Z.}, \bibinfo{author}{Li, C.}, \bibinfo{author}{Xu, J.}, \bibinfo{author}{Liu, Y.}, \bibinfo{year}{2022}.
\newblock \bibinfo{title}{Rgbt salient object detection: A large-scale dataset and benchmark}.
\newblock \bibinfo{journal}{IEEE Transactions on Multimedia} \bibinfo{volume}{25}, \bibinfo{pages}{4163--4176}.
\bibitem[{Wang et~al.(2025)Wang, Tsao, Wang, Wang, Feng, Tian and Poria}]{wang2025action}
\bibinfo{author}{Wang, J.}, \bibinfo{author}{Tsao, R.}, \bibinfo{author}{Wang, X.}, \bibinfo{author}{Wang, X.}, \bibinfo{author}{Feng, F.}, \bibinfo{author}{Tian, S.}, \bibinfo{author}{Poria, S.}, \bibinfo{year}{2025}.
\newblock \bibinfo{title}{Action-guided prompt tuning for video grounding}.
\newblock \bibinfo{journal}{Information Fusion} \bibinfo{volume}{113}, \bibinfo{pages}{102577}.
\bibitem[{Wang et~al.(2024a)Wang, Lin, Li, Tu and Luo}]{wang2024alignment}
\bibinfo{author}{Wang, K.}, \bibinfo{author}{Lin, D.}, \bibinfo{author}{Li, C.}, \bibinfo{author}{Tu, Z.}, \bibinfo{author}{Luo, B.}, \bibinfo{year}{2024}a.
\newblock \bibinfo{title}{Alignment-free rgbt salient object detection: Semantics-guided asymmetric correlation network and a unified benchmark}.
\newblock \bibinfo{journal}{IEEE Transactions on Multimedia} .
\bibitem[{Wang et~al.(2024b)Wang, Wu, Li, Zhao, Chen, Shi and Tang}]{wang2024structural}
\bibinfo{author}{Wang, X.}, \bibinfo{author}{Wu, W.}, \bibinfo{author}{Li, C.}, \bibinfo{author}{Zhao, Z.}, \bibinfo{author}{Chen, Z.}, \bibinfo{author}{Shi, Y.}, \bibinfo{author}{Tang, J.}, \bibinfo{year}{2024}b.
\newblock \bibinfo{title}{Structural information guided multimodal pre-training for vehicle-centric perception}, in: \bibinfo{booktitle}{Proceedings of the AAAI Conference on Artificial Intelligence}, pp. \bibinfo{pages}{5624--5632}.
\bibitem[{Wang et~al.(2022)Wang, Li, Zheng, He and Tang}]{wang2022interact}
\bibinfo{author}{Wang, Z.}, \bibinfo{author}{Li, C.}, \bibinfo{author}{Zheng, A.}, \bibinfo{author}{He, R.}, \bibinfo{author}{Tang, J.}, \bibinfo{year}{2022}.
\newblock \bibinfo{title}{Interact, embed, and enlarge: Boosting modality-specific representations for multi-modal person re-identification}, in: \bibinfo{booktitle}{Proceedings of the AAAI Conference on Artificial Intelligence}, pp. \bibinfo{pages}{2633--2641}.
\bibitem[{Wang et~al.(2019)Wang, Liu, Li, Sheng, Yan, Wang and Shao}]{wang2019camp}
\bibinfo{author}{Wang, Z.}, \bibinfo{author}{Liu, X.}, \bibinfo{author}{Li, H.}, \bibinfo{author}{Sheng, L.}, \bibinfo{author}{Yan, J.}, \bibinfo{author}{Wang, X.}, \bibinfo{author}{Shao, J.}, \bibinfo{year}{2019}.
\newblock \bibinfo{title}{Camp: Cross-modal adaptive message passing for text-image retrieval}, in: \bibinfo{booktitle}{Proceedings of the IEEE/CVF international conference on computer vision}, pp. \bibinfo{pages}{5764--5773}.
\bibitem[{Xu et~al.(2022)Xu, Zhang, Wei, Lin, Cao, Hu and Bai}]{xu2022simple}
\bibinfo{author}{Xu, M.}, \bibinfo{author}{Zhang, Z.}, \bibinfo{author}{Wei, F.}, \bibinfo{author}{Lin, Y.}, \bibinfo{author}{Cao, Y.}, \bibinfo{author}{Hu, H.}, \bibinfo{author}{Bai, X.}, \bibinfo{year}{2022}.
\newblock \bibinfo{title}{A simple baseline for open-vocabulary semantic segmentation with pre-trained vision-language model}, in: \bibinfo{booktitle}{European Conference on Computer Vision}, \bibinfo{organization}{Springer}. pp. \bibinfo{pages}{736--753}.
\bibitem[{Yang et~al.(2023)Yang, Zhou, Zheng, Wang, Zhu and Wu}]{yang2023towards}
\bibinfo{author}{Yang, S.}, \bibinfo{author}{Zhou, Y.}, \bibinfo{author}{Zheng, Z.}, \bibinfo{author}{Wang, Y.}, \bibinfo{author}{Zhu, L.}, \bibinfo{author}{Wu, Y.}, \bibinfo{year}{2023}.
\newblock \bibinfo{title}{Towards unified text-based person retrieval: A large-scale multi-attribute and language search benchmark}, in: \bibinfo{booktitle}{Proceedings of the 31st ACM International Conference on Multimedia}, pp. \bibinfo{pages}{4492--4501}.
\bibitem[{Ying et~al.(2024)Ying, Xiao, Li, He, Li, Li, Wang, Hu, Xu, Lin et~al.}]{ying2024visible}
\bibinfo{author}{Ying, X.}, \bibinfo{author}{Xiao, C.}, \bibinfo{author}{Li, R.}, \bibinfo{author}{He, X.}, \bibinfo{author}{Li, B.}, \bibinfo{author}{Li, Z.}, \bibinfo{author}{Wang, Y.}, \bibinfo{author}{Hu, M.}, \bibinfo{author}{Xu, Q.}, \bibinfo{author}{Lin, Z.}, et~al., \bibinfo{year}{2024}.
\newblock \bibinfo{title}{Visible-thermal tiny object detection: A benchmark dataset and baselines}.
\newblock \bibinfo{journal}{arXiv preprint arXiv:2406.14482} .
\bibitem[{Yuan et~al.(2022)Yuan, Wang and Wei}]{yuan2022translation}
\bibinfo{author}{Yuan, M.}, \bibinfo{author}{Wang, Y.}, \bibinfo{author}{Wei, X.}, \bibinfo{year}{2022}.
\newblock \bibinfo{title}{Translation, scale and rotation: cross-modal alignment meets rgb-infrared vehicle detection}, in: \bibinfo{booktitle}{European Conference on Computer Vision}, \bibinfo{organization}{Springer}. pp. \bibinfo{pages}{509--525}.
\bibitem[{Zeng et~al.(2023)Zeng, Zhang, Li, Wang, Zhang and Zhou}]{zeng2023x}
\bibinfo{author}{Zeng, Y.}, \bibinfo{author}{Zhang, X.}, \bibinfo{author}{Li, H.}, \bibinfo{author}{Wang, J.}, \bibinfo{author}{Zhang, J.}, \bibinfo{author}{Zhou, W.}, \bibinfo{year}{2023}.
\newblock \bibinfo{title}{$x^2$-vlm: All-in-one pre-trained model for vision-language tasks}.
\newblock \bibinfo{journal}{IEEE transactions on pattern analysis and machine intelligence} \bibinfo{volume}{46}, \bibinfo{pages}{3156--3168}.
\bibitem[{Zhang et~al.(2025)Zhang, Feng, Wang, Lu and Mei}]{zhang2025transformer}
\bibinfo{author}{Zhang, X.}, \bibinfo{author}{Feng, Y.}, \bibinfo{author}{Wang, N.}, \bibinfo{author}{Lu, G.}, \bibinfo{author}{Mei, S.}, \bibinfo{year}{2025}.
\newblock \bibinfo{title}{Transformer-based person detection in paired rgb-t aerial images with vtsar dataset}.
\newblock \bibinfo{journal}{IEEE Journal of Selected Topics in Applied Earth Observations and Remote Sensing} .
\bibitem[{Zhang et~al.(2023)Zhang, Xu, Yang, He, Yu, Yu and Xia}]{zhang2023drone}
\bibinfo{author}{Zhang, Y.}, \bibinfo{author}{Xu, C.}, \bibinfo{author}{Yang, W.}, \bibinfo{author}{He, G.}, \bibinfo{author}{Yu, H.}, \bibinfo{author}{Yu, L.}, \bibinfo{author}{Xia, G.S.}, \bibinfo{year}{2023}.
\newblock \bibinfo{title}{Drone-based rgbt tiny person detection}.
\newblock \bibinfo{journal}{ISPRS Journal of Photogrammetry and Remote Sensing} \bibinfo{volume}{204}, \bibinfo{pages}{61--76}.
\bibitem[{Zheng et~al.(2021)Zheng, Wang, Chen, Li and Tang}]{zheng2021robust}
\bibinfo{author}{Zheng, A.}, \bibinfo{author}{Wang, Z.}, \bibinfo{author}{Chen, Z.}, \bibinfo{author}{Li, C.}, \bibinfo{author}{Tang, J.}, \bibinfo{year}{2021}.
\newblock \bibinfo{title}{Robust multi-modality person re-identification}, in: \bibinfo{booktitle}{Proceedings of the AAAI Conference on Artificial Intelligence}, pp. \bibinfo{pages}{3529--3537}.
\bibitem[{Zhou et~al.(2022)Zhou, Dong, Xu and Qian}]{zhou2022edge}
\bibinfo{author}{Zhou, W.}, \bibinfo{author}{Dong, S.}, \bibinfo{author}{Xu, C.}, \bibinfo{author}{Qian, Y.}, \bibinfo{year}{2022}.
\newblock \bibinfo{title}{Edge-aware guidance fusion network for rgb--thermal scene parsing}, in: \bibinfo{booktitle}{Proceedings of the AAAI conference on artificial intelligence}, pp. \bibinfo{pages}{3571--3579}.
\bibitem[{Zuo et~al.(2024a)Zuo, Hong, Zhang, Yu, Zhou, Gao, Sang and Wang}]{zuo2024plip}
\bibinfo{author}{Zuo, J.}, \bibinfo{author}{Hong, J.}, \bibinfo{author}{Zhang, F.}, \bibinfo{author}{Yu, C.}, \bibinfo{author}{Zhou, H.}, \bibinfo{author}{Gao, C.}, \bibinfo{author}{Sang, N.}, \bibinfo{author}{Wang, J.}, \bibinfo{year}{2024}a.
\newblock \bibinfo{title}{Plip: Language-image pre-training for person representation learning}, in: \bibinfo{booktitle}{The Thirty-eighth Annual Conference on Neural Information Processing Systems}.
\bibitem[{Zuo et~al.(2024b)Zuo, Zhou, Nie, Zhang, Guo, Sang, Wang and Gao}]{zuo2024ufinebench}
\bibinfo{author}{Zuo, J.}, \bibinfo{author}{Zhou, H.}, \bibinfo{author}{Nie, Y.}, \bibinfo{author}{Zhang, F.}, \bibinfo{author}{Guo, T.}, \bibinfo{author}{Sang, N.}, \bibinfo{author}{Wang, Y.}, \bibinfo{author}{Gao, C.}, \bibinfo{year}{2024}b.
\newblock \bibinfo{title}{Ufinebench: Towards text-based person retrieval with ultra-fine granularity}, in: \bibinfo{booktitle}{Proceedings of the IEEE/CVF Conference on Computer Vision and Pattern Recognition}, pp. \bibinfo{pages}{22010--22019}.

\end{thebibliography}

\end{document}